\documentclass[10pt]{article} % For LaTeX2e
\usepackage[accepted]{tmlr}
\usepackage{amsmath,amsfonts,bm}

\def\eqref#1{equation~\ref{#1}}
\def\1{\bm{1}}

\def\vh{{\bm{h}}}

\def\vm{{\bm{m}}}

\def\vu{{\bm{u}}}

\def\vx{{\bm{x}}}
\def\vy{{\bm{y}}}
\def\vz{{\bm{z}}}

\def\evepsilon{{\epsilon}}

\def\mH{{\bm{H}}}

\def\mK{{\bm{K}}}

\def\mM{{\bm{M}}}

\def\mU{{\bm{U}}}
\def\mV{{\bm{V}}}
\def\mW{{\bm{W}}}

\def\mY{{\bm{Y}}}

\DeclareMathAlphabet{\mathsfit}{\encodingdefault}{\sfdefault}{m}{sl}
\SetMathAlphabet{\mathsfit}{bold}{\encodingdefault}{\sfdefault}{bx}{n}

\usepackage{hyperref}
\usepackage{url}

\usepackage{xspace}
\usepackage{graphicx}
\usepackage{subfigure}
\usepackage{subcaption}
\usepackage{booktabs}
\usepackage{multirow}
\usepackage{mathtools}

\newcommand{\tb}{\textsc{TextBoost}\xspace}
\newcommand{\tbpp}{\textsc{TextBoost}\texttt{++}\xspace}

\newcommand{\vstar}{\texttt{V$^\star$}\xspace}

\title{Boosting Text Encoder for \\Personalized Text-to-Image Generation}

\author{\name NaHyeon Park\textsuperscript{*} \email julia19@kaist.ac.kr \\
      \addr KAIST
      \AND
      \name Kunhee Kim\textsuperscript{*} \email kunhee.kim@kaist.ac.kr \\
      \addr KAIST
      \AND
      \name Hyunjung Shim \email kateshim@kaist.ac.kr\\
      \addr KAIST}
\def\month{05}  % Insert correct month for camera-ready version
\def\year{2026} % Insert correct year for camera-ready version
\def\openreview{\url{https://openreview.net/forum?id=hiZzk1nHuV}} % Insert correct link to OpenReview for camera-ready version

\begin{document}

\maketitle

\begingroup
\renewcommand\thefootnote{}\NoHyper\footnote{*Equal contribution.}\endNoHyper
\addtocounter{footnote}{-1}
\endgroup

\begin{abstract}
In this paper, we introduce \tb, an efficient one-shot personalization approach for text-to-image diffusion models. Traditional personalization methods typically involve fine-tuning extensive portions of the model, leading to substantial storage requirements and slow convergence. In contrast, we propose selectively fine-tuning only the text encoder, significantly improving computational and storage efficiency. To preserve the original semantic integrity, we develop a novel causality-preserving adaptation mechanism. Additionally, lightweight adapters are employed to locally refine text embeddings immediately before their interaction with cross-attention layers, greatly enhancing the expressiveness of text embeddings with minimal computational overhead. Empirical evaluations across diverse concepts demonstrate that \tb achieves faster convergence and substantially reduces storage demands by minimizing the number of trainable parameters. Furthermore, \tb maintains comparable subject fidelity, superior text fidelity, and greater generation diversity compared to existing methods. We show that our proposed method offers an efficient, scalable, and practically applicable solution for high-quality text-to-image personalization, particularly beneficial in resource-constrained environments.
\end{abstract}

%%%%%%%%%% INTRODUCTION %%%%%%%%%%
\section{Introduction}
Recent advancements in text-to-image (T2I) diffusion models have significantly expanded the creative potential of AI-driven image synthesis, enabling high-fidelity images to be generated from natural language prompts \citep{ramesh_zero-shot_2021, rombach_high-resolution_2022, balaji_ediff-i_2022, nichol_glide_2023, chen2024pixartalpha, podell_sdxl_2024, esser_scaling_2024}.
Despite these advances, standard text prompts often lack the specificity required to capture fine details or unique artistic styles, leading to ambiguities in the generated images. 
To address this, personalization techniques have emerged as a critical research direction, aiming to capture a specific concept (\vstar) within the token space so that the generated images faithfully reflect user-defined content \citep{gal_image_2023, ruiz_dreambooth_2023}.

While existing personalization methods show promising results, they often suffer from computational and storage inefficiencies. 
For instance, DreamBooth \citep{ruiz_dreambooth_2023} fine-tunes all parameters of the U-Net and saves a distinct model for each concept, which demands substantial storage overhead. 
Moreover, current methods often require multiple reference images (usually three to five), limiting their practicality in real-world scenarios where users can only provide a single reference image. 
These issues underscore the need for efficient approaches to personalization—both in terms of parameter usage and training data requirements.

In this paper, we propose an efficient one-shot personalization method by focusing on an underexplored yet pivotal component of the T2I model: the text encoder. 
Typically, personalization approaches fix the text encoder and fine-tune components within the U-Net, such as cross-attention modules \citep{kumari_multi-concept_2023, chen_disenbooth_2024}. However, our initial empirical analysis suggests that when all parameters of the T2I model are fine-tuned, the text encoder undergoes the largest changes, indicating significant potential benefits if adapted properly.

Building on this insight, we introduce \textbf{\tb}, a novel method that selectively fine-tunes the CLIP text encoder to achieve efficient one-shot personalization. 
% A key challenge lies in preserving the rich semantic representations already encoded in CLIP.
% We address this with a \textit{causality-preserved adaptation} (CPA) mechanism, which ensures that embeddings for tokens preceding \vstar tokens remain unaltered while allowing new concept embeddings to be learned. 
A key challenge is to fine-tune the text encoder without disrupting its learned semantic space, particularly its auto-regressive causality: the property where a token's output embedding is conditioned only on itself and prior tokens in the sequence. To address this, we introduce Causality-Preserved Adaptation (CPA), a mechanism specifically designed to ensure that during the learning of a new concept token (\vstar), the output embeddings of all tokens preceding \vstar remain unchanged. This preserves the original contextual understanding of the sequence leading up to the new concept.
Concretely, CPA formulates the adapted output as a mixture of original and fine-tuned embeddings, preventing the distortion of established semantic structures of the original text encoder.

To further enhance fine-grained concept control, we draw inspiration from hierarchical representations in GAN architectures \citep{karras_style-based_2019, karras_analyzing_2020, richardson_encoding_2021,tov_designing_2021} and introduce lightweight adapters that locally refine text embeddings before they interact with each cross-attention layer. 
Unlike token-space methods that rely on multiple forward passes \citep{voynov_p_2023, alaluf_neural_2023}, our design requires only a single pass through these adapters, substantially improving computational efficiency and runtime.

We validate \tb across various datasets, comparing it to popular personalization methods such as DreamBooth \citep{ruiz_dreambooth_2023}, Textual Inversion \citep{gal_image_2023}, LoRA \citep{hu_lora_2022}, and Custom Diffusion \citep{kumari_multi-concept_2023}. 
Our results demonstrate that \tb achieves faster convergence, reduced storage overhead, high subject and text fidelity, thus offering an improved balance between efficiency and generative quality.

In summary, our key contributions are as follows:
\begin{itemize} 
    \item \textbf{Fine-tuning text encoders for T2I personalization}: We present a novel approach for fine-tuning text encoders specifically designed for text-to-image (T2I) personalization, a previously unexplored direction.
    \item \textbf{Causality-preserved adaptation (CPA)}: We introduce a causality-preserved adaptation strategy that maintains the integrity of the text encoder's original semantic properties during adaptation, thereby preserving the causal relationships embedded within the semantic space.
    \item \textbf{Extended text embedding}: We propose extending the embedding space using lightweight adapters, facilitating more precise and fine-grained adaptation of concepts through enhancements in the cross-attention mechanism.
    \item \textbf{Efficient personalization}: Our method significantly improves training efficiency and reduces storage requirements, enabling high-quality T2I personalization suitable for deployment in resource-constrained real-world applications.
\end{itemize}

%%%%%%%%%% RELATED WORK %%%%%%%%%%
\section{Related Work}

\subsection{Personalized text-to-image generation}
Recent advancements in text-to-image (T2I) generation have increasingly emphasized personalization, allowing models to capture user-specific concepts more accurately. Textual Inversion~\citep{gal_image_2023} introduced encoding personalized information into special learned tokens (\vstar tokens). Building upon this, \citep{voynov_p_2023} proposed a more expressive representation with layer-wise prompt embeddings for detailed concept inversion across different model layers. \citep{alaluf_neural_2023} further extended this paradigm with a timestep-wise inversion approach, offering greater flexibility but at the cost of increased computational overhead and longer training and inference times.
Parallel efforts using fine-tuning-based methods, such as DreamBooth~\citep{ruiz_dreambooth_2023}, achieve superior fidelity by updating significant portions of the diffusion model, yet incur substantial computational and storage demands. To address these limitations, techniques like Custom Diffusion~\citep{kumari_multi-concept_2023} and OFT~\citep{qiu_controlling_2023} explored partial model fine-tuning, reducing computational load but still requiring modifications to the image generation component.
Encoder-focused adaptations, including ELITE~\citep{wei_elite_2023} and BLIP-Diffusion~\citep{li_blip-diffusion_2023}, present another pathway to reduce computational overhead by primarily updating encoder components, thus minimizing trainable parameters while preserving personalized output quality. However, these encoder-based methods typically depend on extensive pre-training on large-scale datasets.

\subsection{Parameter-efficient fine-tuning}
Parameter-efficient fine-tuning (PEFT) methods have recently emerged as crucial techniques for effectively training large foundation models with limited resources.
Adapter-based methods~\citep{houlsby_parameter-efficient_2019} insert lightweight layers into pretrained models, enabling efficient adaptation. 
\citet{hu_lora_2022} introduced a simpler and highly efficient technique by inserting low-rank adaptation (LoRA) layers in parallel to existing weights, allowing the modifications to be merged back into the original model parameters after training. 
Due to its efficiency and effectiveness, LoRA has become popular in personalized T2I diffusion tasks, inspired subsequent studies \citep{qiu_controlling_2023, ruiz_hyperdreambooth_2024, shah_ziplora_2024} and the basic component of recent personalization methods \citep{chen_disenbooth_2024, lee_direct_2024}. 
Similarly, StyleDrop utilized adapters for transformer-based diffusion models, demonstrating effective customization capabilities~\citep{sohn_styledrop_2023}.
Our proposed method also leverages techniques from the PEFT domain, such as adapters~\citep{houlsby_parameter-efficient_2019} and LoRA~\citep{hu_lora_2022}. 
However, unlike previous approaches primarily aimed at optimizing downstream task performance, our method emphasizes preserving the original capabilities of pretrained models by adaptively fine-tuning parameters in a causality-preserved manner.

\section{Motivation}
%------------------------------------------
\begin{figure}[t!]
    \centering
    \includegraphics[width=0.4\linewidth]{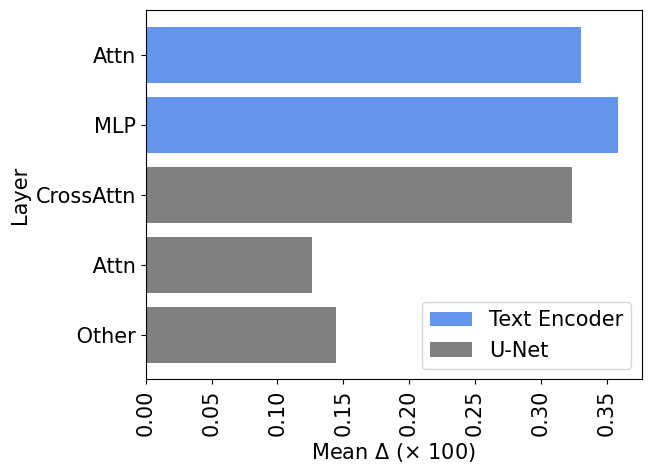}
    \caption{\textbf{Change in weights of different layers during fine-tuning of a T2I model.} The mean weight change in the text encoder layers is notably larger compared to that of the U-Net parameters. This suggests that the text encoder plays a pivotal role in the personalization task.}
    \label{fig:delta}
\end{figure}
%------------------------------------------

Unlike previous works, which typically fine-tune either the concept token \vstar or specific components of the U-Net, we begin by systematically re-evaluating which parts of the text-to-image (T2I) diffusion model are most critical to fine-tuning for effective personalization.

\citet{kumari_multi-concept_2023} proposed an efficient approach by fine-tuning only the cross-attention layers of the U-Net, motivated by observing substantial parameter changes within these layers during personalization~\citep{li_few-shot_2020}. 
Specifically, they targeted the key and value projection layers. 
However, their analysis did not consider potential changes in the text encoder parameters, a critical component responsible for generating embeddings fed into the cross-attention layers.

Given that key and value projections directly depend on text embeddings, we hypothesized that the text encoder might also significantly influence personalization outcomes. 
To rigorously test this hypothesis, we extended the investigation from \citep{kumari_multi-concept_2023} to include both the U-Net and the text encoder. 
We quantified parameter changes during fine-tuning using the following measure:
\begin{equation}
\setlength\abovedisplayskip{1pt}
\setlength\belowdisplayskip{1pt}
    \Delta = \frac{||\hat{\theta} - \theta||}{||\theta||},
    \label{eq:delta}
\end{equation}
where $\theta$ and $\hat{\theta}$ denote the original and fine-tuned weights.

Consistent with previous findings~\citep{kumari_multi-concept_2023}, we observed significant changes in the U-Net's cross-attention layer parameters. 
Surprisingly, as illustrated in Fig.~\ref{fig:delta}, the text encoder parameters exhibited even more substantial adjustments. 
This finding strongly highlights the previously overlooked importance of the text encoder in personalized T2I generation, aligning with recent insights emphasizing its central role~\citep{saharia_photorealistic_2022, chen_enhancing_2024, li_textcraftor_2024}.

To the best of our knowledge, this study is the first systematic exploration specifically targeting fine-tuning the text encoder for personalized text-to-image generation.

%%%%%%%%%% METHOD %%%%%%%%%%
\section{Boosting text encoder tuning}

Building upon insights from the preceding section, we introduce \tb, a novel approach for personalizing text-to-image diffusion models.
% We begin by providing a concise overview of these models (Section~\ref{sec:prelim}). Subsequently, we detail our adaptation methodology, which incorporates the causal properties of the CLIP text encoder (Section~\ref{sec:cpa}).
% Finally, we present a method for expanding the CLIP text representation space into a localized embedding space (Section~\ref{sec:expand}).

\begin{figure}[t!]
    \centering
    \includegraphics[width=0.55\linewidth]{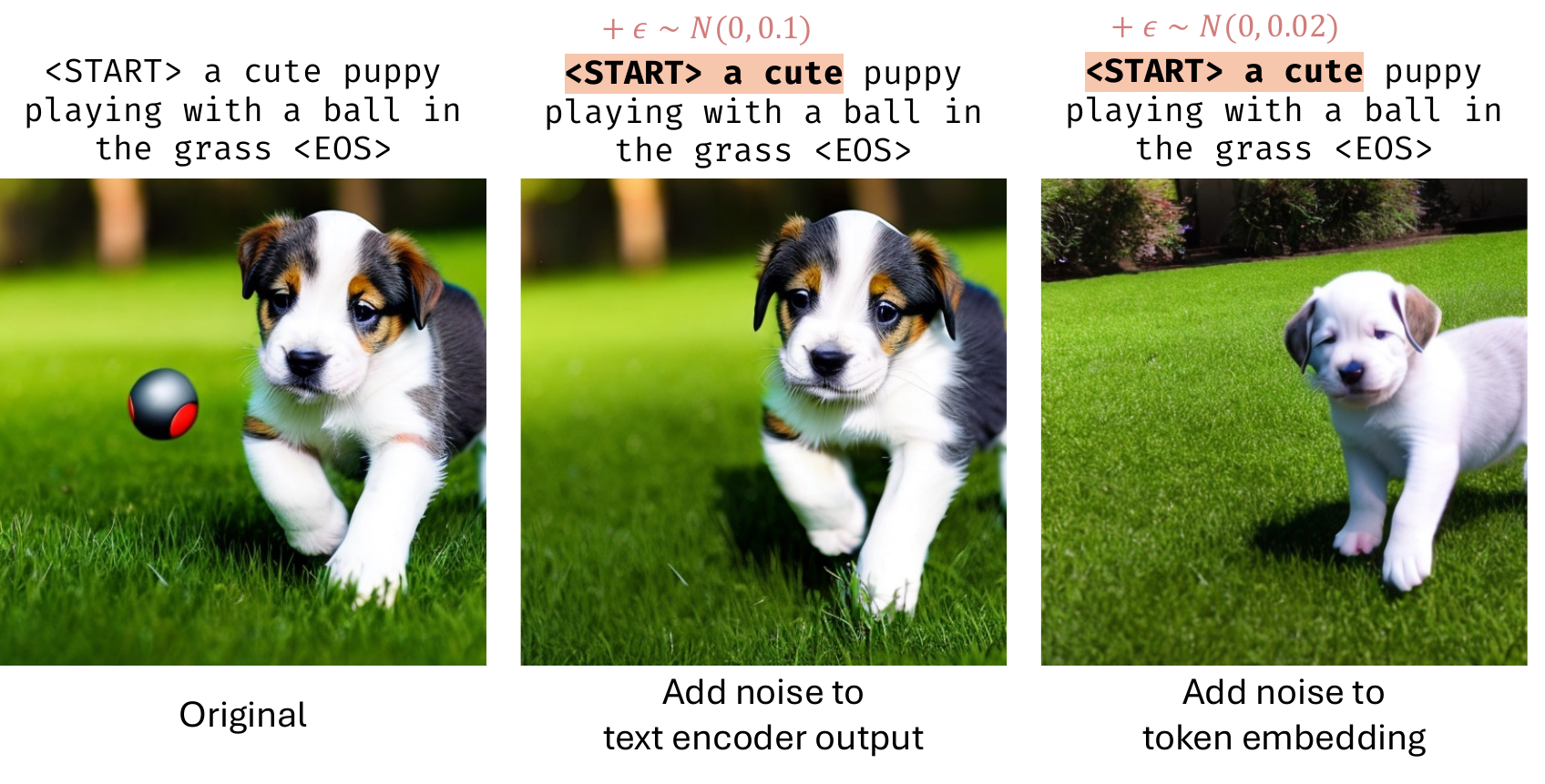}
    \caption{\textbf{Effect of perturbations on early token embeddings during text-to-image generation.} We compare the original generation (left) with images produced after adding small Gaussian noise to the text encoder's output embeddings (middle) and directly to the token embeddings (right). Adding even slight noise significantly affects generated image quality, highlighting the sensitivity and importance of preserving embeddings for early tokens in text-to-image generation.}
    \label{fig:challenge}
\end{figure}

\subsection{Preliminaries: Text-to-image diffusion models}
\label{sec:prelim}
Diffusion models~\citep{sohl-dickstein_deep_2015} have recently become the most widely adopted generative models for text-to-image generation~\citep{saharia_photorealistic_2022, ramesh_hierarchical_2022, rombach_high-resolution_2022}.
% They learn a model distribution $p_\theta(\vx_0)$ that approximates the data distribution $q(\vx_0)$.
These models aim to closely approximate the original data distribution $q(\vx_0)$ with $p_\theta(\vx_0)$.
% Let $p_\theta(\vx_{0:T})$ denote the reverse process, a Markov chain with learned Gaussian transitions, so that $p_\theta(\vx_0)=\int p_\theta(\vx_{0:T})\, d\vx_{1:T}$.
Here, $p_\theta(\vx_0) \coloneqq \int p_\theta (\vx_{0:T})$, where $p_\theta(\vx_{0:T})$ is termed as the reverse process being Markov chain with learned Gaussian transitions.
% The forward process $q(\vx_{1:T}\mid \vx_0)$ gradually adds noise to the original sample $\vx_0$, yielding $\vx_t=\sqrt{\alpha_t}\vx_0+\sqrt{1-\alpha_t}\evepsilon$, and factorizes as $q(\vx_{1:T}\mid \vx_0)=\prod_{t=1}^{T} q(\vx_t\mid \vx_{t-1})$.
The approximate posterior $q(\vx_{1:T} | \vx_0)$ is termed a forward process, of which the noise is gradually added to the original data point $\vx_0$ as $\vx_t = \sqrt{\alpha_t} \vx_0 + \sqrt{1- \alpha_t} \evepsilon$, can be expressed as $q(\vx_{1:T} | \vx_0) \coloneqq \prod_{t=1}^{T} q(\vx_t | \vx_{t-1})$.
The training objective is formulated with variational bound on negative log-likelihood:
\begin{equation}
    \mathbb{E}[- \log p_\theta(\vx_0)] \le \mathbb{E}_q[- \log \frac{p_\theta (\vx_{0:T})}{q(\vx_{1:T} | \vx_0)}] \coloneqq \mathcal{L},
    \label{eq:nll}
\end{equation}
which is further simplified into the following objective:
\begin{equation}
    \mathcal{L}_{simple}(\theta) \coloneqq \mathbb{E}_{t, \vx_0, \evepsilon}[|| \evepsilon - \evepsilon_{\theta}(\vx_t, t)||^{2}_{2}].
    \label{eq:lsimple}
\end{equation}

Text-to-image (T2I) diffusion models~\citep{rombach_high-resolution_2022} take text condition $\textbf{c}$ as additional input, where the text prompt $y$ is encoded with text encoder $\mathcal{E}_{T}$ as $\textbf{c} = \mathcal{E}_{T}(y)$.
These models usually operate in latent space, encoding an image $\vx_0$ into a latent $\vz_0 = \mathcal{E}_{I}(\vx_0)$.
The forward diffusion process is then applied in latent space, yielding $\vz_t = \sqrt{\alpha_t}\vz_0 + \sqrt{1-\alpha_t}\evepsilon$.
The training objective hence takes an additional $\textbf{c}$ and minimizes the following loss:
\begin{equation}
    \mathbb{E}_{\vz_0, t, y, \evepsilon} \left[ \| \evepsilon - \evepsilon_{\theta}(\vz_t, t, \textbf{c}) \|^{2}_{2} \right].
    \label{eq:t2i}
\end{equation}

While our approach generalizes to various T2I models, we primarily employ Stable Diffusion~\citep{rombach_high-resolution_2022}, due to its extensive usage and open accessibility. Stable Diffusion adopts a U-Net architecture~\citep{ronneberger_u-net_2015} for its denoising diffusion process. 
To guide the image generation, text embeddings are injected through cross-attention layers, which dynamically integrate these embeddings into the U-Net at each diffusion step.

Given the critical role of the CLIP text encoder~\citep{radford_learning_2021} in widely-used T2I diffusion models such as Stable Diffusion (SD)~\citep{rombach_high-resolution_2022} and Stable Diffusion XL (SDXL)~\citep{podell_sdxl_2024}, we specifically targeted the CLIP text encoder for effective personalization in our method.

\begin{figure*}[t]
    \centering
    % --- Subfigure (a) ---
    \begin{minipage}[b]{0.18\linewidth} % b: 하단 기준 정렬
        % \centering
        \subfigure[Causal attention mask]{
        \raisebox{7mm}{ % 이 수치를 키우면 더 올라갑니다 (예: 40mm, 50mm)
            \includegraphics[width=\linewidth]{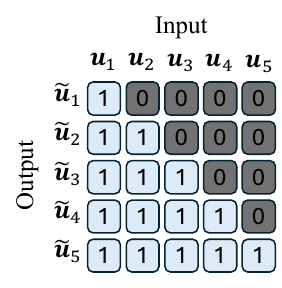}
        }
        \label{fig:causal_mask}
    }
    \end{minipage}
    \hspace{2mm} % 이미지 사이 간격 조절
    % --- Subfigure (b) ---
    \begin{minipage}[b]{0.5\linewidth}
        \centering
        \subfigure[Causality of CLIP text encoder]{
            \includegraphics[width=\linewidth]{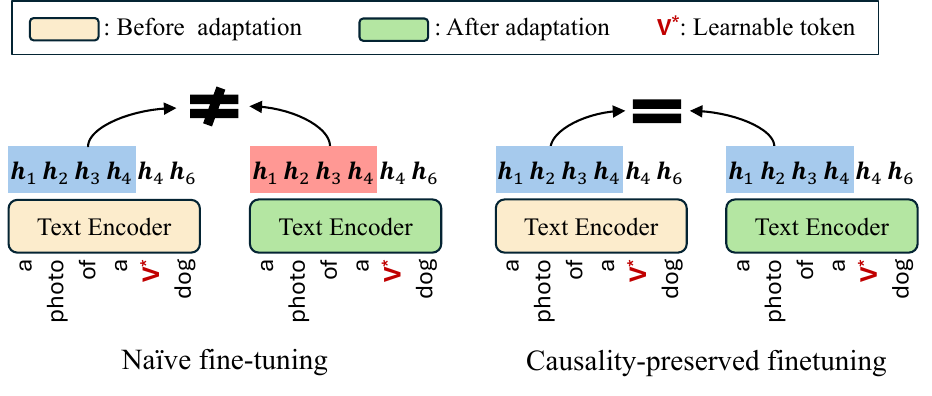}
            \label{fig:causal}
        }
    \end{minipage}
    \hspace{2mm}
    % --- Subfigure (c) ---
    \begin{minipage}[b]{0.25\linewidth}
        \centering
        \subfigure[Causality-preserved adapter]{
            \includegraphics[width=\linewidth]{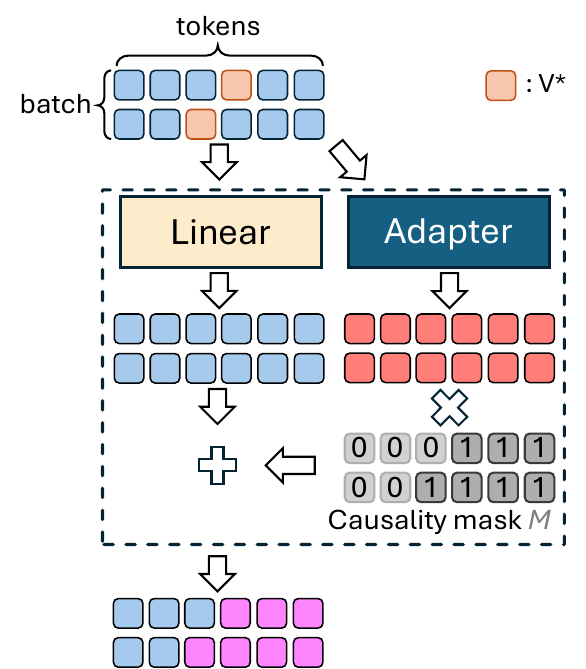}
            \label{fig:cpa}
        }
    \end{minipage}

    \caption{\textbf{(a) Causal attention mask.} The CLIP text encoder employs causal masking techniques, where the output embedding $\vh_i$ corresponding to the input token $\vy_i$ can only attend to preceding tokens. \textbf{(b) Causality of CLIP text encoder.} If this causality is not properly accounted for during fine-tuning (e.g., through naive fine-tuning), the embeddings of tokens preceding the new concept token \vstar can be unintentionally altered, leading to undesired changes in their representation. \textbf{(c) Our causality-preserved adapter.} We attach adapters in parallel to the linear layers, and incorporate a causality mask to ensure that the fine-tuned adapter influences only the new concept token \vstar and the subsequent tokens, thereby maintaining the causality structure.}
    \label{fig:causal_all}
\end{figure*}

\subsection{Causality-preserved adaptation}
\label{sec:cpa}
\subsubsection{Challenges in fine-tuning CLIP text encoder}
The CLIP text encoder, like many language models, is a decoder-only transformer~\citep{radford_learning_2021} that employs causal masking (see Fig.~\ref{fig:causal_mask}). This architectural feature enforces an auto-regressive processing flow: the output embedding $\vh_i$ for an input token $\vy_i$ is generated based solely on $\vy_i$ and all preceding tokens ($\vy_1, \dots, \vy_{i-1}$). It is not influenced by any subsequent tokens ($\vy_j$ where $j > i$). This inherent sequential causality is fundamental to how the encoder builds contextual representations.

When fine-tuning the encoder to learn a new concept token \vstar, a naive update to the model's weights can inadvertently alter the output embeddings of tokens preceding \vstar (as illustrated in Fig.~\ref{fig:causal} where $h_1$ to $h_4$ change). This happens because weight changes, even if optimized for \vstar, propagate throughout the model. Such alterations disrupt the established semantic interpretations of the initial part of the text sequence (see Fig.~\ref{fig:challenge}), which can degrade the model's ability to understand prompts and accurately depict the concept. Our goal is to adapt the model for \vstar while strictly preserving the original output embeddings of all tokens $\vh_i$ that appear before \vstar in any given prompt.

\subsubsection{Causality-preserved adapter}
One straightforward solution would involve fully fine-tuning the text encoder and subsequently overwriting embeddings preceding \vstar with their original values. 
This approach, however, requires two full forward passes, leading to inefficiencies and a failure to maintain layer-wise causality.
Instead, we propose a more efficient causality-preserving adapter (CPA) inspired by parameter-efficient fine-tuning methods~\citep{houlsby_parameter-efficient_2019, hu_lora_2022} from large language models (LLMs). 
We attach adapters in parallel to selected linear layers within the Transformer.
Let $\mU=[\vu_1; \dots; \vu_n] \in \mathbb{R}^{n \times d_{\mathrm{in}}}$ denote the matrix of token representations entering an adapted linear layer.
The adapted output is computed as
\begin{equation}
    \label{eq5}
    \widetilde{\mU} = \mU\mW
    +
    \mathrm{Diag}(\vm) \mathcal{A}_{\mathrm{CPA}}(\mU),
\end{equation}
where $\mW \in \mathbb{R}^{d_\mathrm{in}\times d_\mathrm{out}}$ is the frozen pretrained weight matrix and
$\mathcal{A}_\mathrm{CPA}(\mU) = \mU\mW_{down}\mW_{up}$ represents the adapter with LoRA-style down-projection $\mW_{down} \in \mathbb{R}^{d_{in} \times r}$ and up-projection $\mW_{up} \in \mathbb{R}^{r \times d_{out}}$.
For a tokenized prompt $\mY=[\vy_1;\dots;\vy_n]$, with the special token \vstar located at position $i^\star$, the causal mask vector $\vm=[m_1,\dots,m_n]^\top \in \{0,1\}^n$ is defined as
\begin{equation}
    m_i =
    \begin{cases}
        1, & \text{if } i \ge i^\star \text{ and } \vy_i \text{ is not a padding token},\\
        0, & \text{otherwise.}
    \end{cases}
\end{equation}
For a batch of $B$ prompts, these vectors can be stacked row-wise into a batched mask matrix $\mM_{\mathrm{CPA}} \in \{0,1\}^{B\times n}$, which is the form illustrated in Fig.~\ref{fig:cpa}.
This mask ensures that the adapter affects only \vstar and the subsequent non-padding tokens.
For all tokens with $i < i^\star$, we have $m_i=0$, so the $i$-th row of the adapter term in Eq.~\ref{eq5} is zero and the adapted output reduces to the pretrained output.
This preserves the original encoder representations for the initial part of the sequence and maintains the auto-regressive causality up to \vstar (see Fig.~\ref{fig:cpa}). We call this overall strategy Causality-Preserved Adaptation (CPA).

Based on empirical findings and our ablation studies (see Fig.~\ref{fig:ablation}), we incorporated adapters into specific layers for optimal performance. Notably, our approach is different from some previous conventions that primarily target self-attention parameters with adapters\footnote{\url{https://github.com/huggingface/diffusers/blob/main/examples/dreambooth/train_dreambooth_lora.py}}. Instead, our ablations indicated that focusing on other components yielded better results for our personalization task. Specifically, within Multi-Layer Perceptron (MLP) blocks of the text encoder, we add adapters to the second feed-forward layer (\texttt{fc2}).
The rank parameter $r$ determines the adapter's bottleneck dimension, controlling both parameter efficiency and expressivity; unless otherwise stated, we adopt $r=1$ as our default configuration.

% ------------------------------------------
\begin{figure}[t]
    \centering
    \includegraphics[width=0.65\linewidth]{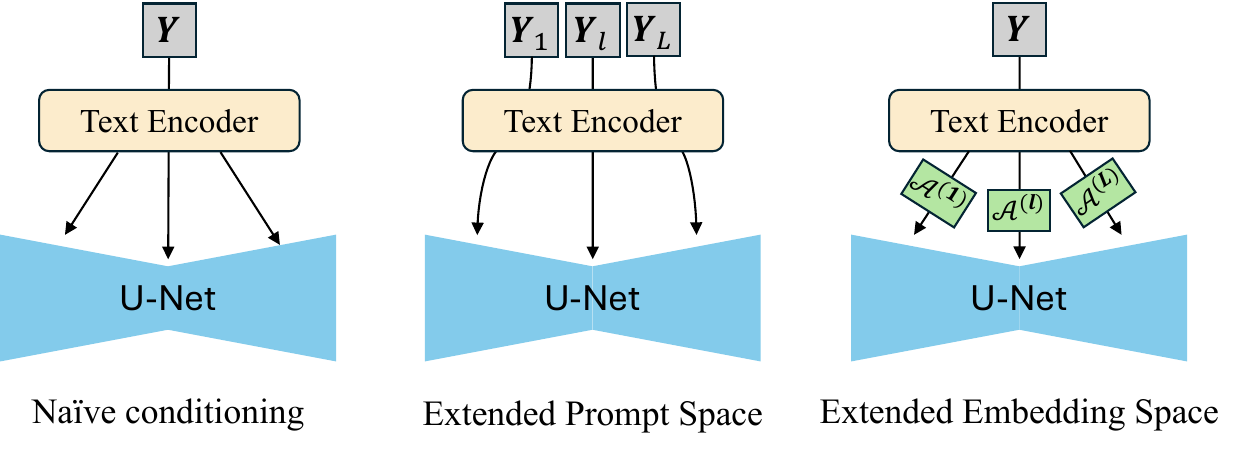}
    \caption{\textbf{Extended embedding space.} Unlike previous methods that operate in the prompt space, our approach introduces adapters in the embedding space, resulting in an extended embedding space. This design allows for more flexible integration of the fine-tuned text embedding from the text encoder into the image generation module. Here, $\mY$ represents the input text prompt, and $\mathcal{A}^{(\ell)}$ denotes the adapter for cross-attention layer $\ell$.}
    \label{fig:extend}
\end{figure}
% ------------------------------------------

\subsection{Extended textual embedding}
\label{sec:expand}
%% [TODO] Custom Diffusion의 formulation을 다르게 하면
%% W'_k x = (W_k + delta) x = W_kx + delta x
%% 따라서 custom diffusion은 causality를 고려하지 않은 hierarchical representation.
%% Custom diffusion과 무엇이 다르냐는 질문에 대해 유사한 점을 인정하고 서술. 오히려 다른 점을 강조해서 미리 defence할 수 있도록 할 것.

Extended latent representations have proven effective in achieving more precise inversion in generative adversarial networks~\citep{goodfellow_generative_2014, karras_style-based_2019}, as demonstrated by methods like Image2StyleGAN~\citep{abdal_image2stylegan_2019} or e4e~\citep{tov_designing_2021}.
Motivated by these advances, we propose to extend the CLIP text encoder into an extended representation space.

Let a tokenized prompt $\mY = [\vy_1;\dots;\vy_n]$ be encoded by the CLIP text encoder $\mathcal{E}_T$, producing final hidden states
\begin{equation}
\mH = \mathcal{E}_T(\mY) \in \mathbb{R}^{n \times d}.
\end{equation}
In standard text-to-image diffusion models, this single embedding matrix $\mH$ is shared across all cross-attention layers of the U-Net, where it is projected to keys and values.

We define \emph{extended textual embedding} as a layer-wise conditioning scheme that replaces the single text-encoder hidden state $H$ with a set of per-cross-attention embeddings $\{\mH^{(\ell)}\}_{\ell=1}^{L_{\mathrm{ca}}}$ where $L_{\mathrm{ca}}$ denotes the number of cross-attention layers in the U-Net.
\begin{equation}
\mH^{(\ell)} = \mH + \mathrm{Diag}(\vm)\mathcal{A}_\ell(\mH),
\end{equation}
where $\mathcal{A}_\ell(\cdot)$ denotes a lightweight adapter specific to cross-attention layer $\ell$, and $\vm \in \{0,1\}^{n}$ is the causality mask defined in Section~\ref{sec:cpa}.

Each $\mH^{(\ell)}$ is injected into cross-attention block $\ell$ \emph{before} its key and value projection layers:
\begin{equation}
\mK_\ell = \mH^{(\ell)} \mW_\ell^K, \qquad
\mV_\ell = \mH^{(\ell)} \mW_\ell^V.
\end{equation}
Importantly, the projection matrices are unchanged; instead, each cross-attention layer receives its own adapted textual embedding. This effectively extends the conditioning space from a single embedding $\mH$ to a collection $(\mH^{(1)}, \dots, \mH^{(L_\mathrm{ca})})$, increasing representational flexibility while maintaining architectural simplicity.

Crucially, to preserve causality, embeddings corresponding to tokens positioned before the novel concept token $\vstar$ remain unchanged due to the masking term $\mathrm{Diag}(\vm)$.

As illustrated in Fig.~\ref{fig:extend}, we apply causality-preserved adaptation (CPA) to refine text embeddings separately for each cross-attention layer.
Unlike token-space methods such as XTI~\citep{voynov_p_2023} or NeTI~\citep{alaluf_neural_2023}, which require multiple forward passes through the text encoder, our approach performs only one forward pass through the text encoder followed by lightweight adapter transformations to construct $\{\mH^{(\ell)}\}_{\ell=1}^{L_\mathrm{ca}}$.
This design significantly reduces computational overhead and improves efficiency compared to hierarchical prompt-embedding techniques.

We refer to the text-encoder fine-tuning approach with CPA as \tb, and the extended version incorporating both CPA and extended textual embedding as \tbpp.

\section{Experiments}

\subsection{Experimental setup}
\paragraph{Dataset.}
We employed the benchmark introduced by \cite{ruiz_dreambooth_2023}, which consists of 30 subjects and 25 text prompts, with each subject having 4 to 6 associated images.
For style experiments, we use the reference images provided by StyleDrop~\citep{sohn_styledrop_2023}.
It is important to note that we trained all of the models using a single reference image.
Following DreamBooth, we generated 4 images per prompt (3,000 images in total) for evaluation. 

\paragraph{Evaluation metrics.}
We evaluate image-text alignment, which measures how well the image reflects the user prompt, and subject fidelity, which assesses how accurately the image represents the characteristics of the given concept.
Regarding image-text alignment, while prior works commonly used CLIP score, recent findings by~\citep{lin_evaluating_2024} indicate that the VQA score better reflects human perception. Therefore, we opted for the VQA score for a more accurate evaluation.
Subject fidelity was assessed by computing the pairwise cosine similarity between generated and reference image embeddings, following the methodology of ~\citep{ruiz_dreambooth_2023}. We employed \texttt{DINOv2:ViT/L-14} from DINOv2~\citep{oquab_dinov2_2024} as our feature extractor, aligning with recent approaches~\citep{lee_direct_2024, jang_decordecomposition_2024}.
We used the foreground of the validation images as suggested by \citep{kim_selectively_2024}. To obtain foreground images, we employed GroundingDINO~\citep{liu_grounding_2024} and GroundedSAM~\citep{ren_grounded_2024}.
Since these scores do not evaluate the extent of overfitting, memorization of reference images can result in inflated scores.
Considering this, we assessed the inter-similarity between generated images using the DINOv2~\citep{oquab_dinov2_2024} embedding.
Finally, to assess the practical applicability of our method, we conducted a user study involving 50 participants on Amazon Mechanical Turk to evaluate user preferences.

% ------------------------------------------
\begin{table}[t]
\centering
\caption{\textbf{Quantitative comparison on Stable Diffusion v2.1-base.} We measure VQA scores~\citep{lin_evaluating_2024} for image-text fidelity. Pairwise DINOv2~\citep{oquab_dinov2_2024} feature cosine similarity is evaluated with the reference images that are not used for training. For practicality, we compare the number of parameters.}
\resizebox{0.8\linewidth}{!}{
    \begin{tabular}{l   rrr   r}
    \toprule
    \textbf{Methods} & VQA $\uparrow$ & DINOv2 $\uparrow$ & Diversity $\uparrow$ & \# Params $\downarrow$  \\ \midrule
    Textual Inversion~\citep{gal_image_2023} & 0.475 & 0.467 & \textbf{0.408} & 0.00M \\
    NeTI~\citep{alaluf_neural_2023} & 0.602 & 0.536 & 0.279 & 0.83M \\\midrule
    DreamBooth~\citep{ruiz_dreambooth_2023} & 0.553 & 0.587 & 0.164 & 865.91M \\
    DreamBooth-LoRA~\citep{hu_lora_2022} & 0.528 & 0.589 & 0.190 & 0.83M \\
    Custom Diffusion~\citep{kumari_multi-concept_2023} & 0.452 & \textbf{0.632} & 0.136 & 25.56M \\
    DisenBooth~\citep{chen_disenbooth_2024} & \underline{0.675} & 0.486 & 0.305 & 2.93M \\\midrule
    \tb & \textbf{0.680} & 0.560 & \underline{0.307} & \textbf{0.12M} \\
    \tbpp & 0.607 & \underline{0.598} & 0.229 & \textbf{0.15M} \\ \bottomrule
    \end{tabular}
}
\label{tab:clipscore}
\end{table}
% ------------------------------------------
%------------------------------------------
\begin{table}[t]
    \centering
    \caption{\textbf{Quantitative comparison with encoder-based methods.} We compare \tb with encoder-based methods, which require pre-training. The results highlight the superior performance of our method over the baselines in terms of subject similarity and text fidelity.}
    \resizebox{0.5\linewidth}{!}{
    \begin{tabular}{l c rrr}
        \toprule
        \textbf{Methods} & T2I Model & VQA $\uparrow$ & DINOv2 $\uparrow$  \\ \midrule
        ELITE & \multirow{2}{*}{SD1.4} & 0.430 & 0.489 \\
        \tbpp & & \textbf{0.558} & \textbf{0.564} \\ \midrule
        BLIP Diffusion & \multirow{2}{*}{SD1.5} & \textbf{0.650} & 0.531 \\
        \tbpp & & 0.577 & \textbf{0.575} \\ \bottomrule
    \end{tabular}
    }
    \label{tab:oneshot}
\end{table}
%------------------------------------------
%--------------------------------------------------------
\begin{figure}[t]
\centering
\includegraphics[width=0.4\linewidth]{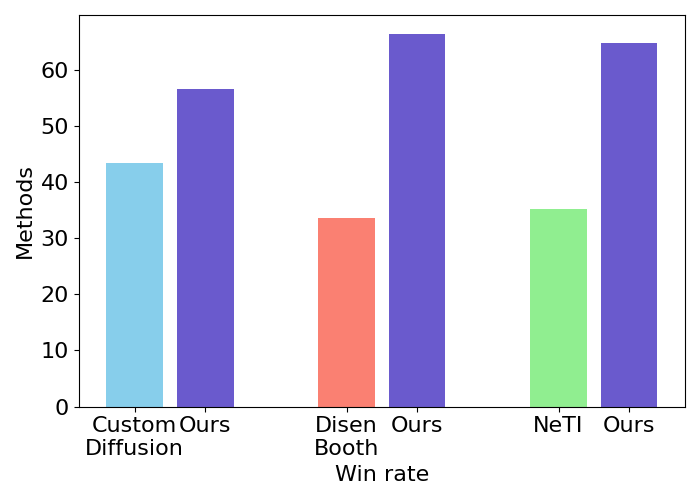}
\caption{\textbf{User study.} We conducted a large-scale user-preference survey with 50 participants on Amazon Mechanical Turk. We asked each participant 30 questions to choose an image that best resembles the given subject and aligns with the text prompt at the same time (1,500 responses in total).}
\label{fig:userstudy}
\end{figure}
%--------------------------------------------------------

% --- Figure 1: Qualitative Comparison (독립) ---
\begin{figure}[t]
    \centering
    % 첫 번째 이미지 섹션
    \begin{minipage}{\linewidth}
        \centering
        \includegraphics[width=\linewidth]{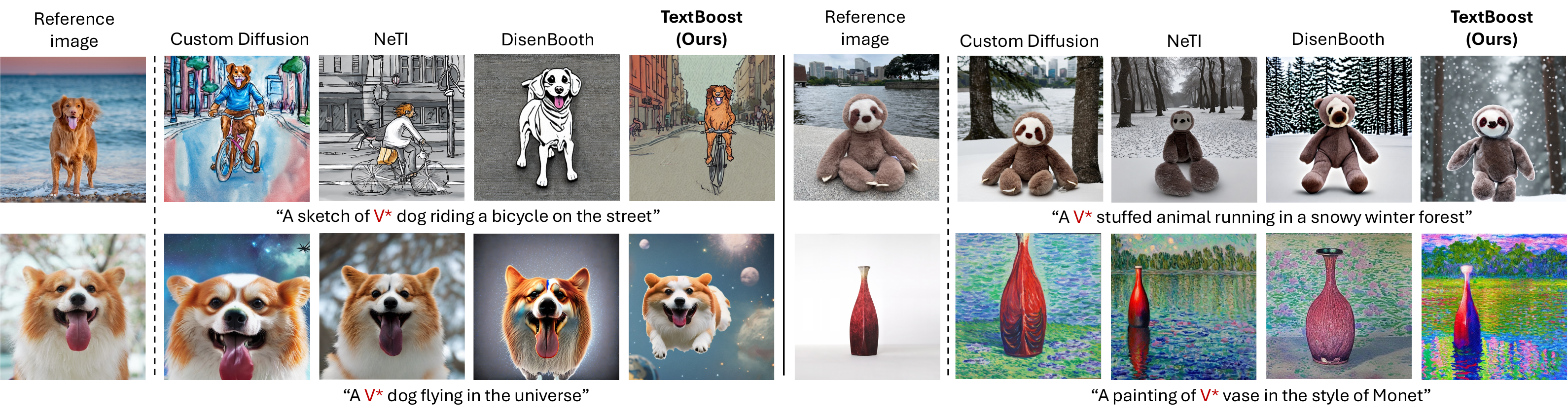}
        \caption{{\bf Qualitative comparison.} We compare images generated by a diverse set of methods, utilizing various types of text prompts across different subjects. It is important to note that all models were trained using a \emph{single} reference image, positioned at the leftmost of each comparison. The results demonstrate that our \tb method can accurately generate images that adhere to the given prompt, maintaining high subject fidelity, even when compared to more computationally intensive methods with a large number of parameters.}
        \label{fig:quali}
    \end{minipage}

    \vspace{2em} % 두 이미지 사이의 간격 조절

    % 두 번째 이미지 섹션
    \begin{minipage}{\linewidth}
        \centering
        \includegraphics[width=\linewidth]{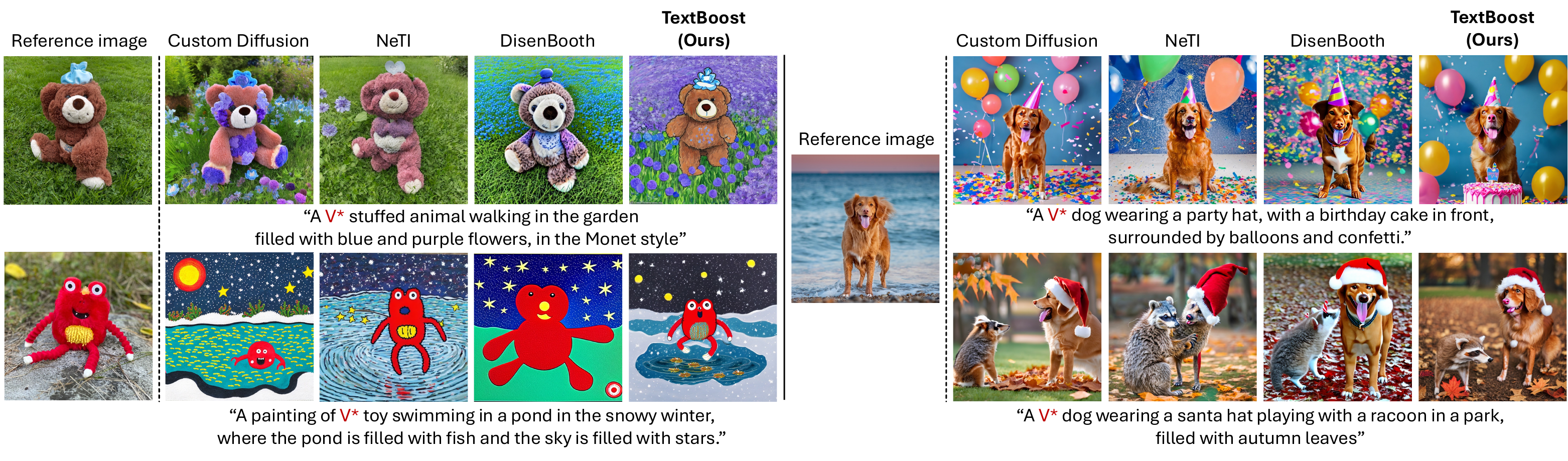}
        \caption{{\bf Comprehensive captions.} To demonstrate the effectiveness of \tb in handling detailed captions, we generate images from long and creative descriptions that vary in attributes, backgrounds, and styles. Compared to other baseline methods, \tb exhibits superior performance in accurately reflecting these comprehensive captions, maintaining high subject fidelity.}
        \label{fig:quali-comprehensive}
    \end{minipage}
\end{figure}

\paragraph{Baselines.}
We compare our approach with recent personalization methods: DreamBooth~\citep{ruiz_dreambooth_2023}, DreamBooth-LoRA~\citep{hu_lora_2022}, Textual Inversion \citep{gal_image_2023}, Custom Diffusion \citep{kumari_multi-concept_2023}, NeTI~\citep{alaluf_neural_2023} and DisenBooth~\citep{chen_disenbooth_2024}.
We also compare our method with zero-shot personalization methods \citep{wei_elite_2023,li_blip-diffusion_2023}, which require pre-training beforehand. 

\paragraph{Implementation details.}
We utilized the \texttt{imagenet small} template for the training prompt, following \citep{gal_image_2023}. For the identifier, we employed templates of the form \vstar \texttt{[class]}, similar to Custom Diffusion. For example, our prompt was constructed as, ``a photo of a nice \vstar \texttt{dog}''. We also implemented weak augmentation, though significantly less aggressive than that used in Custom Diffusion~\citep{kumari_multi-concept_2023}. Furthermore, we did not modify the loss computation region using masking. For the T2I model, we used various versions of Stable Diffusion~\citep{rombach_high-resolution_2022} model, integrating the text encoder from the CLIP~\citep{radford_learning_2021} or OpenCLIP~\citep{cherti_reproducible_2023}.
Our adapter is trained using the AdamW optimizer \citep{loshchilov_decoupled_2019} with a learning rate of 1e-4 for text encoder and 1e-3 for {\texttt V$^*$} token for $100$ steps. For \tbpp, we trained for $80$ steps.
All experiments were conducted with batch size 8 on a single NVIDIA A6000 GPU.

\subsection{Quantitative results}
The quantitative comparison is presented in Tab.~\ref{tab:clipscore}. Our method performs on par with existing approaches while using significantly fewer parameters. \tb achieves a notably high VQA score, demonstrating strong performance. Regarding subject fidelity, our method consistently produces results that are comparable to or exceed those of other methods, suggesting that with proper tuning of the text encoder, effective personalization can be achieved without sacrificing performance. Additionally, our method excels in generating a diverse range of output images, as indicated by the high diversity score, providing end users with diverse options.

We also compare our method, \tb, with ELITE and BLIP Diffusion in Tab.~\ref{tab:oneshot}. As shown in the table, \tb achieves comparable or even superior image and text fidelity compared to these encoder-based methods, which typically require extensive pre-training. This highlights a key advantage of our approach, as it does not require pre-training or subject data, enabling rapid and efficient fine-tuning during test time.

\paragraph{Computational efficiency.}
We emphasize the practical advantages of our method, which demands significantly fewer trainable parameters, as detailed in Tab.~\ref{tab:clipscore}. In this regard, the fact that our method achieves performance comparable to existing approaches is a significant advantage. Moreover, personalization introduces storage challenges, as the fine-tuned model must be saved for each concept. Consequently, the number of parameters directly correlates with storage requirements, as all parameters related to the fine-tuned components of the T2I model must be stored. While Textual Inversion offers notable storage efficiency due to its focus on fine-tuning text embeddings, its limited performance hinders practical application, as evidenced by the low image fidelity and image-text alignment scores shown in the table. The compact size of our model allows it to be seamlessly stored in environments with limited storage capacity, such as the cloud or portable storage devices.

\paragraph{User study.}
In real-world scenarios, users need to select the output image that both satisfies (1) subject fidelity and (2) alignment between the text prompt and the generated image.
To evaluate existing methods and understand user preferences, we conducted a large-scale user study. 
Participants were tasked with selecting the best image from multiple options generated by different methods. We employed diverse subjects and text prompts sourced from \citep{ruiz_dreambooth_2023} (templates in Appendix~\ref{sec:app:user_study}). 
To ensure fairness, random seeds were fixed, and we computed the win-rate between our method and previous works.
Through Amazon Mechanical Turk, 50 participants completed 30 questions each, yielding a total of 1,500 responses. 
As indicated in Fig.~\ref{fig:userstudy}, our method showed higher win-rates compared to all of the existing methods, demonstrating its superior ability to meet user demands for subject fidelity and text-image alignment in practical settings.

% --- Figure 3 & 4: Stylization & Diversity (한 줄에 합침) ---
% 첫 번째 이미지: Stylization
\begin{figure}[t]
    \centering
    \includegraphics[width=0.7\linewidth]{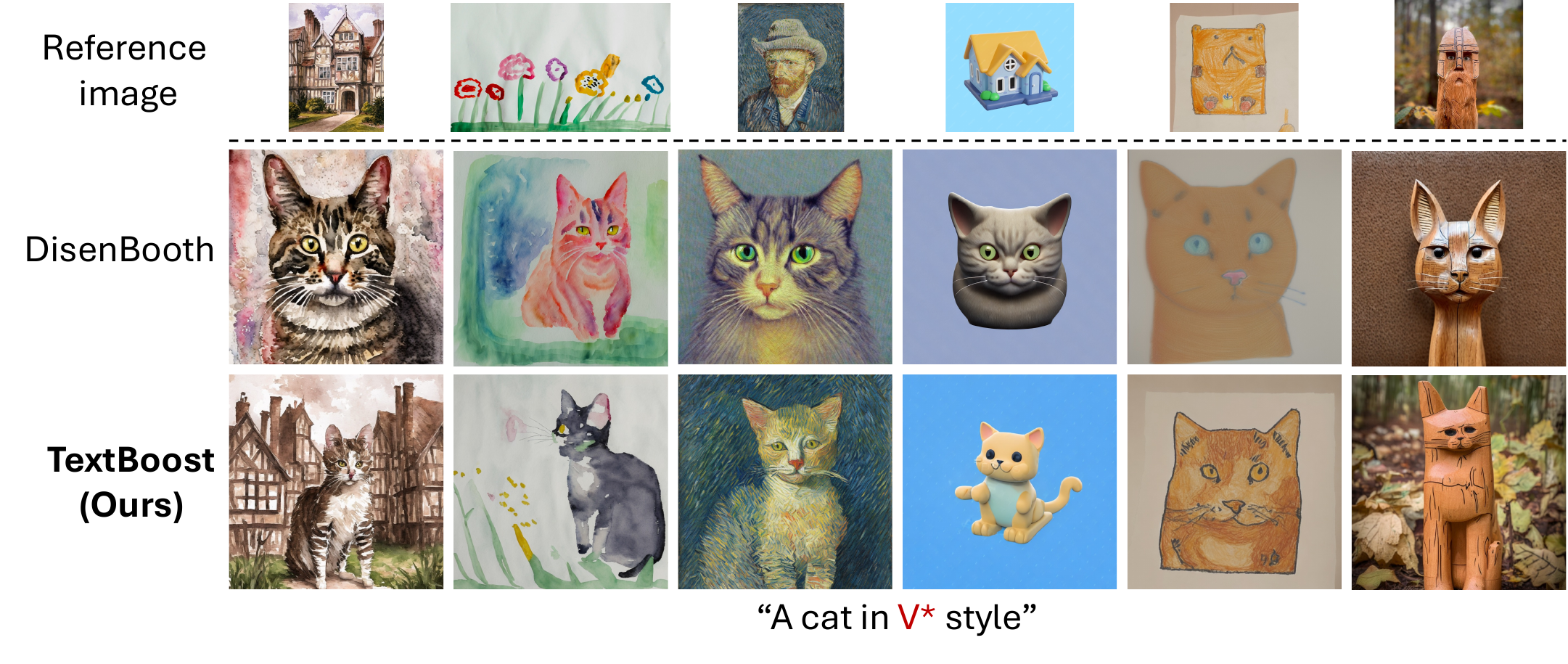}
    \caption{\textbf{Results on stylization.} The output generated by \tb effectively captures the reference style while maintaining key details such as texture, color patterns, and artistic nuances. In comparison to the recent method, DisenBooth, our \tb demonstrates superior fidelity in reproducing these stylistic features.}
    \label{fig:style}
\end{figure}

% 두 번째 이미지: Diversity
\begin{figure}[t]
    \centering
    \includegraphics[width=0.7\linewidth]{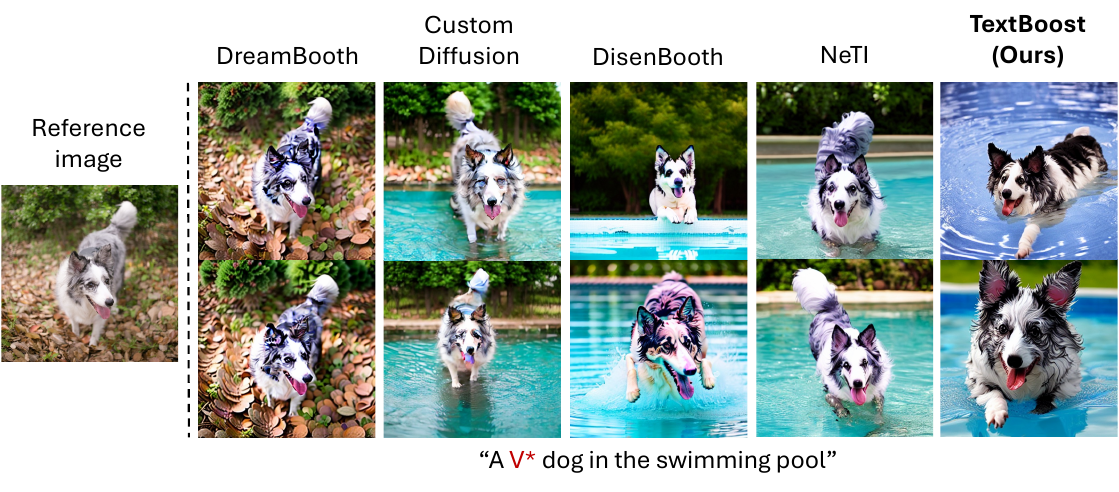}
    \caption{\textbf{Diversity comparison.} We evaluate the inter-similarity of images generated with identical prompts across various baselines. Our \tb consistently generates images with high diversity.}
    \label{fig:diversity}
\end{figure}

\subsection{Qualitative results} 
We conducted a qualitative analysis to compare the baseline models with \tb, and the results are presented in Fig.~\ref{fig:quali} and Fig.~\ref{fig:quali-comprehensive}.

To begin with, Fig.~\ref{fig:quali} illustrates examples generated from creative text prompts. \tb consistently outperforms other methods in accurately reflecting the user's prompt. For instance, in the prompt ``A \vstar dog riding a bicycle on the street'', alternative models exhibit several issues, such as the omission of the bicycle (DisenBooth), the incorrect display of only the bicycle (NeTI), or the occurrence of attribute leakage (e.g., a blue shirt and a street in the image from Custom Diffusion). In contrast, \tb correctly includes all the elements specified in the prompt. More qualitative comparison results can be found in Appendix~\ref{sec:app:comparison}.

We further evaluated \tb with more complex prompts that involve simultaneous modifications to attributes, background, and style, as depicted in Fig.~\ref{fig:quali-comprehensive}. Our findings demonstrate that \tb accurately interprets and generates images based on these comprehensive captions. For example, the term ``walking'' in the top-left prompt posed difficulties for baseline models in capturing the requisite details and style, whereas \tb effectively represented these elements. In the bottom-right example, the inclusion of the word ``raccoon'' caused confusion in other models, resulting in artifacts or misidentification of a dog. However, \tb accurately generated the output of both the dog and raccoon as intended.

These results indicate that \tb enhances prompt alignment and ensures subject accuracy, making it an ideal tool for users seeking greater control over the creative process. Additional results, including experiments on facial datasets, can be found in Appendices~\ref{sec:app:more_results} and \ref{sec:app:face}.

\paragraph{Stylization.}
We also conducted experiments on style personalization using a single reference image. Leveraging the style references from \citep{sohn_styledrop_2023} on Stable Diffusion v2.1, we evaluated our approach in comparison to DisenBooth~\citep{chen_disenbooth_2024}, as alternative methods proved ineffective for this particular task. 
For example, NeTI tends to overfit, resulting in images that fail to capture the intended subject—such as a cat—while only reflecting the style (see Fig.~\ref{fig:neti-style} in Appendix~\ref{sec:app:comparison}). 
As demonstrated in Fig.~\ref{fig:style}, our method effectively learns and applies the style from a single reference image, maintaining fidelity to the content of the text prompt.

\paragraph{Diversity.}
In Tab.~\ref{tab:clipscore}, we measured the inter-similarity of output images using DINOv2, finding that our \tb exhibits greater diversity than other approaches. A visual example is shown in Fig.~\ref{fig:diversity}, where other baselines produce similar images for the same prompts, while our method generates more varied poses and backgrounds. This demonstrates \tb's advantage in generating diverse outputs, highlighting its strong potential for real-world applications. More results can be found in Appendix~\ref{sec:app:more_results}.

\subsection{Ablation study}
%------------------------------------------
\begin{figure*}[t]
    \centering
    % --- Subfigure (a) ---
    \begin{minipage}[b]{0.43\linewidth} % b: 하단 기준 정렬
        \centering
        \subfigure[Effect of CPA]{
        \raisebox{1mm}{ % 이 수치를 키우면 더 올라갑니다 (예: 40mm, 50mm)
            \includegraphics[width=\linewidth]{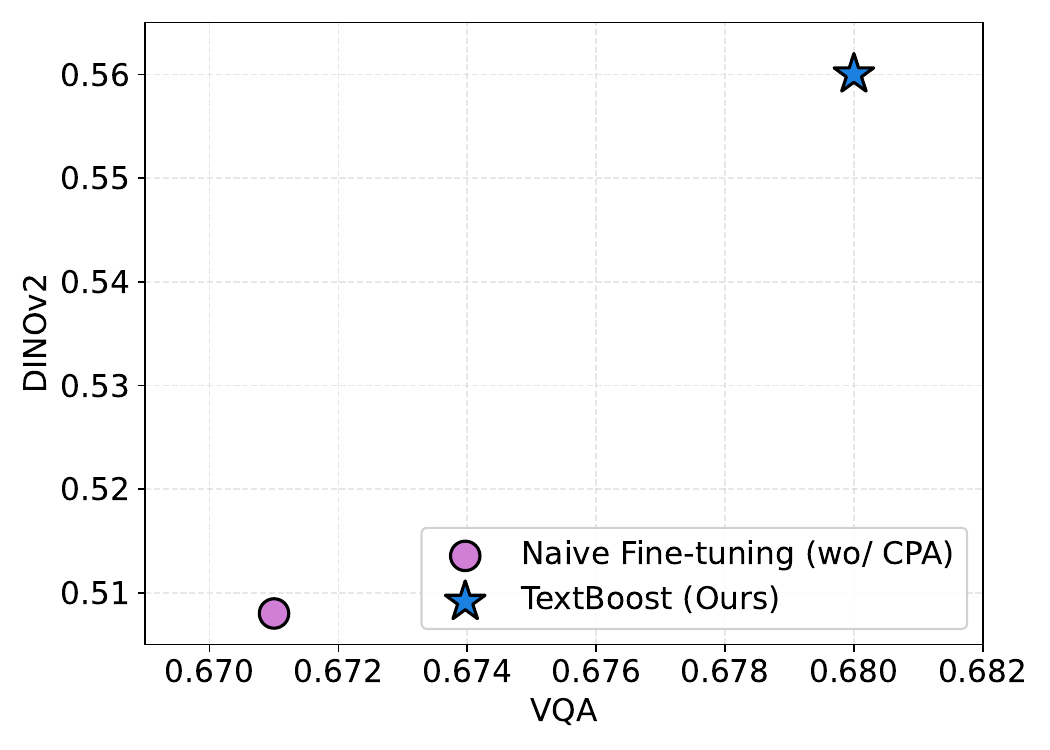}
        }
        \label{fig:ab1}
    }
    \end{minipage}
    \hspace{2mm} % 이미지 사이 간격 조절
    % --- Subfigure (b) ---
    \begin{minipage}[b]{0.54\linewidth}
        \centering
        \subfigure[Effect of adapter placement]{
            \includegraphics[width=\linewidth]{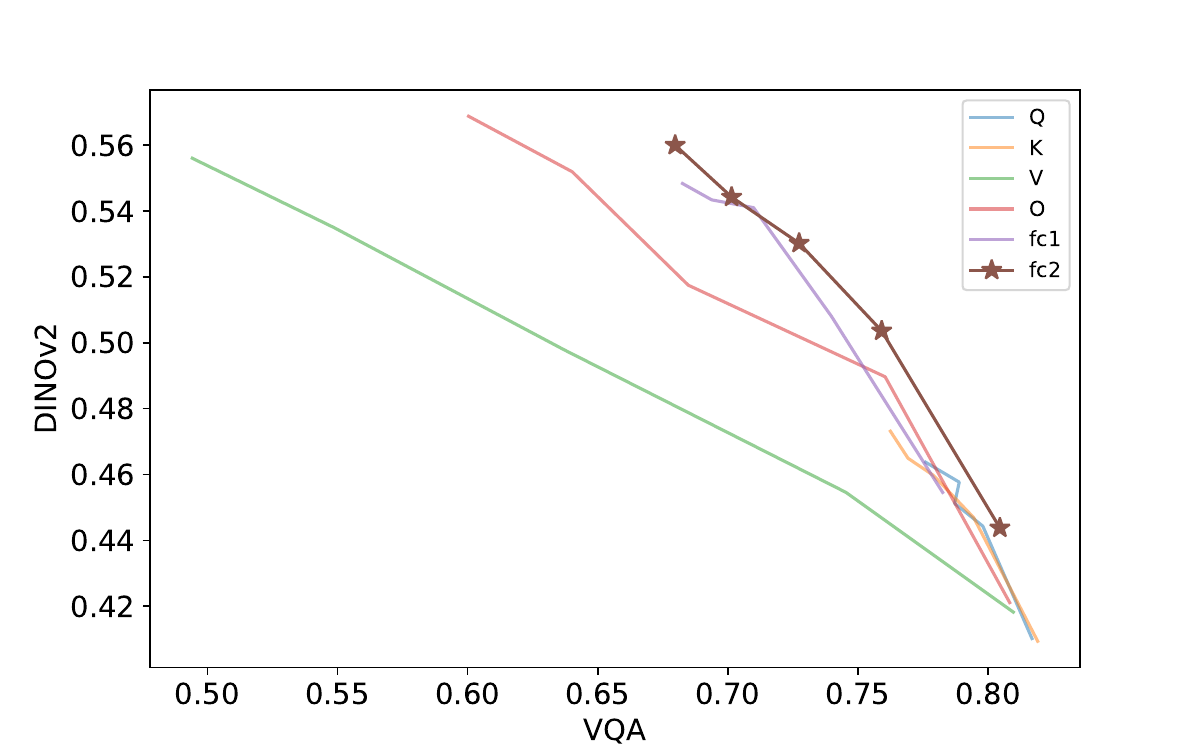}
            \label{fig:ab2}
        }
    \end{minipage}
    \caption{\textbf{Ablation study.} (a) Our proposed CPA results in better image and text fidelity (b) Placing the adapter in \texttt{fc2} shows best performance.}
    \label{fig:ablation}
\end{figure*}
%------------------------------------------------
% \begin{table}[t]
%     \centering
%     \begin{tabular}{l rr}
%     \toprule
%     \textbf{Methods} & \textbf{VQA} & \textbf{DINOv2}
%     \\\midrule\midrule
%     TextBoost++ & 0.610 & 0.596 \\\midrule
%     w/o CPA & \TODO{0.626} & \TODO{0.741} \\
% \bottomrule
%     \end{tabular}
%     \caption{\textbf{Ablation study.} We perform an ablation study to evaluate the individual components of our proposed TextBoost framework. Specifically, CPA refers to causality-preserved adaptation, and expanding denotes the expanding mechanism. The results demonstrate the effectiveness of each component in our method.}
%     \label{tab:abstudy}
% \end{table}

% NOTE: Since we removed TextBoost++, we should use another results.
% \begin{figure}[t]
%     \centering
%     \includegraphics[width=0.7\linewidth]{figures/abalation.pdf}
%     \caption{\textbf{Ablation study.} We perform an ablation study to evaluate the individual components of our proposed TextBoost framework. Specifically, CPA refers to causality-preserved adaptation, and expanding denotes the expanding mechanism. The results demonstrate the effectiveness of each component in our method.}
%     \label{fig:abstudy}
% \end{figure}

Finally, we conduct an ablation study on our causality-preserved adaptation (CPA) and the adapter placement.

The ablation results show the effectiveness of our CPA; its removal from \tb markedly lowered VQA and DINOv2 scores, demonstrating its significant contribution (see Fig.~\ref{fig:ab1}). Note that Tab.~\ref{tab:clipscore} details our extended embedding space's effect.

While parameter-efficient fine-tuning (PEFT) often targets self-attention modules (e.g., LoRA), systematic studies of optimal PEFT targets are lacking. 
We thus explored six linear layers in transformer blocks: self-attention query (\texttt{Q}), key (\texttt{K}), value (\texttt{V}), output (\texttt{O}), and MLP layers (\texttt{fc1}, \texttt{fc2}). 
As shown in Fig.~\ref{fig:ab2}, MLP layer updates results in better text-image alignment compared to adapting self-attention such as \texttt{K} and \texttt{V} projections, verifying our integration of adapters into the MLP's \texttt{fc2} layer.
% To assess the effectiveness of the causality-preserved adaptation of our method, we conducted an ablation study, with the results presented in Fig.~\ref{fig:abstudy}. The findings show that the removal of CPA from TextBoost leads to a noticeable decline in both the VQA score and the DINOv2 score, thereby demonstrating the contribution of casuality-preserved adaptation to the overall performance.

% \textcolor{blue}{
% Here, we note that the computational cost can be further reduced by attaching the adapter only to specific blocks. To examine this, we selectively apply LoRA to early, middle, and late blocks. Nevertheless, applying LoRA to all blocks still achieves the highest overall performance, while maintaining significantly lower computational cost compared to existing methods.
% }

%------------------------------------------
\begin{figure}[t]
    \centering
    \includegraphics[width=0.7\linewidth]{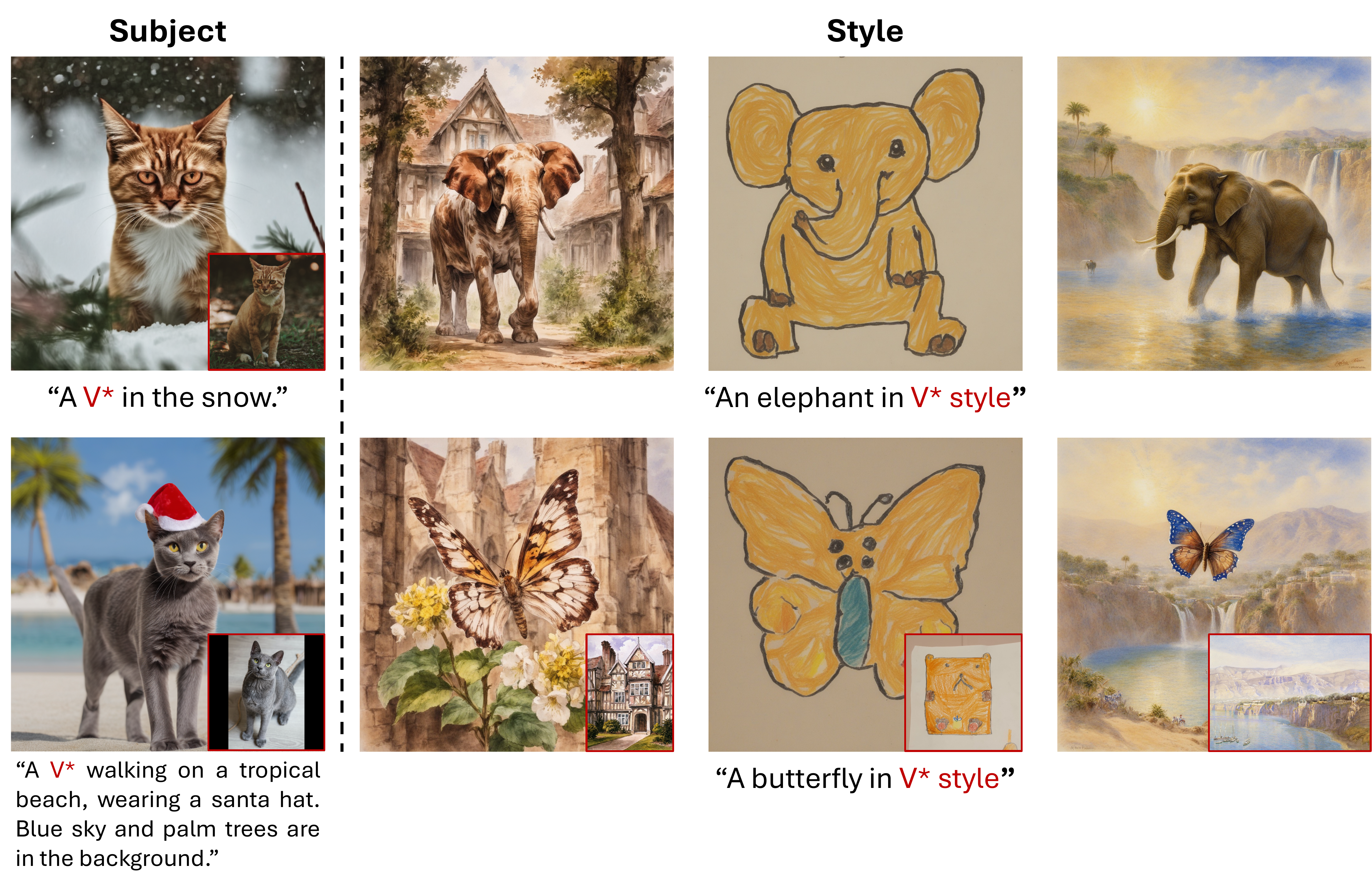}
    \caption{\textbf{Results on SDXL.} The results show that our proposed \tb efficiently scales to larger T2I models, which require multiple text encoders while maintaining a high level of efficiency (a small number of trainable parameters and low storage requirements), which is practical, especially for large T2I models.}
    \label{fig:sdxl}
\end{figure}
%------------------------------------------
\subsection{Results on SDXL}
In widely used models such as Stable Diffusion v1.5 and v2.1, a single CLIP text encoder is used. However, the larger SDXL model has two text encoders: CLIP ViT-L and OpenCLIP ViT-bigG~\citep{cherti_reproducible_2023}.
To test the scalability of our approach, we tested \tb on larger models.
Specifically, we fine-tune only the OpenCLIP ViT-bigG encoder of SDXL while keeping the CLIP ViT-L encoder frozen. This choice is motivated by the ablation results in Tab.~\ref{app:tab:sdxl}, which suggest that adapting a single text encoder provides a favorable trade-off between personalization performance and computational efficiency.
The results, shown in Fig.~\ref{fig:sdxl}, demonstrate that our \tb method effectively captures subject and style characteristics across a variety of prompts. In our GPU setup, methods like DreamBooth and Custom Diffusion require too many parameters, making them less efficient for personalization due to memory limits. On the other hand, our method uses far fewer parameters, enabling faster training and better T2I personalization within SDXL. This shows that our approach is more practical and better suited for resource-limited environments, such as real-world applications, research, or edge devices.

%%%%%%%%%% DISCUSSION %%%%%%%%%%
\section{Discussion \& Conclusion}

The role of text encoders in large-scale text-to-image (T2I) models has attracted growing attention in recent years. Prior works~\citep{saharia_photorealistic_2022, chen2024enhancing, li2024textcraftor} has shown that strengthening the text side of the model can yield substantial gains in generation quality and text-image alignment.  Despite these advances, the potential of text encoder adaptation for personalization has remained largely underexplored, with most existing methods focusing instead on the image generation module or on learning concept-specific token representations.

In this work, we revisit this design choice and show that the text encoder is not merely an auxiliary conditioning component, but a highly effective locus for personalization. Through \tb, we demonstrate that carefully adapting the text encoder can achieve personalization performance comparable to updating the image module, while maintaining strong diversity and offering a substantially more efficient parameterization. These findings suggest that high-quality one-shot personalization does not necessarily require extensive modification of the diffusion backbone; rather, significant gains can be achieved by selectively and structurally adapting the text-side representation space.

More broadly, our results highlight the importance of understanding how personalization is encoded and propagated through the conditioning pathway of T2I models. We believe this perspective opens up new opportunities for designing lightweight, scalable, and practically deployable personalization methods, particularly in resource-constrained settings where storage and training efficiency are critical. Looking forward, we expect that further investigation into text-side adaptation---including richer encoder architectures, more expressive conditioning schemes, and broader concept domains such as faces and artistic styles---will lead to more flexible and reliable personalization systems. We hope \tb\ serves as a step toward this direction and stimulates further research on text-centric approaches to efficient T2I personalization.

\section*{Version note}
A preliminary arXiv version of this work (\texttt{arXiv:2409.08248v1}) used an earlier formulation based on augmentation tokens, knowledge preservation, and SNR-weighted timestep sampling. The present TMLR version substantially revises the method by introducing Causality-Preserved Adaptation (CPA) and the extended textual embedding variant \tbpp, together with updated experiments, ablations, and additional analyses. At the same time, both versions share the same central idea: fine-tuning the text encoder for efficient one-shot text-to-image personalization.

\section*{Acknowledgements}
This work was supported by the Institute for Information \& communications Technology Planning \& Evaluation (IITP) grant funded by the Korea government (MSIT) (RS-2019-II190075, Artificial Intelligence Graduate School Program (KAIST)); this research was supported by the Basic Science Research Program through the National Research Foundation of Korea (NRF) funded by the MSIP (No. RS-2025-00520207); by the IITP grant funded by the Korea government (MSIT) and KEIT grant funded by the Korea government (MOTIE) (No. 2022-0-01045); by the IITP grant funded by the Korea government (MSIT) and KEIT grant funded by the Korea government (MOTIE) (No. 2022-0-00680); and by the Institute of Information \& communications Technology Planning \& Evaluation (IITP) grant funded by the Korea government (MSIT) (No. RS-2024-00457882, National AI Research Lab Project).

%%%%%%%%%% REFERENCES %%%%%%%%%%
\bibliography{main}

@inproceedings{chen_enhancing_2024,
  title={Enhancing diffusion models with text-encoder reinforcement learning},
  author={Chen, Chaofeng and Wang, Annan and Wu, Haoning and Liao, Liang and Sun, Wenxiu and Yan, Qiong and Lin, Weisi},
  booktitle={European Conference on Computer Vision},
  year={2024},
}

@inproceedings{chen2024pixartalpha,
  title={PixArt-\${\textbackslash}alpha\$: Fast Training of Diffusion Transformer for Photorealistic Text-to-Image Synthesis},
  author={Junsong Chen and Jincheng YU and Chongjian GE and Lewei Yao and Enze Xie and Zhongdao Wang and James Kwok and Ping Luo and Huchuan Lu and Zhenguo Li},
  booktitle={International Conference on Learning Representations},
  year={2024},
}

@inproceedings{kim_selectively_2024,
	title = {Selectively {Informative} {Description} can {Reduce} {Undesired} {Embedding} {Entanglements} in {Text}-to-{Image} {Personalization}},
	booktitle = {Proceedings of the {IEEE}/{CVF} {Conference} on {Computer} {Vision} and {Pattern} {Recognition}},
	author = {Kim, Jimyeong and Park, Jungwon and Rhee, Wonjong},
	year = {2024},
}

@inproceedings{liu_grounding_2024,
	title = {Grounding {DINO}: {Marrying} {DINO} with {Grounded} {Pre}-{Training} for {Open}-{Set} {Object} {Detection}},
	booktitle = {European {Conference} on {Computer} {Vision}},
	author = {Liu, Shilong and Zeng, Zhaoyang and Ren, Tianhe and Li, Feng and Zhang, Hao and Yang, Jie and Jiang, Qing and Li, Chunyuan and Yang, Jianwei and Su, Hang and Zhu, Jun and Zhang, Lei},
	year = {2024},
}

@misc{ren_grounded_2024,
	title = {Grounded {SAM}: {Assembling} {Open}-{World} {Models} for {Diverse} {Visual} {Tasks}},
    author={Tianhe Ren and Shilong Liu and Ailing Zeng and Jing Lin and Kunchang Li and He Cao and Jiayu Chen and Xinyu Huang and Yukang Chen and Feng Yan and Zhaoyang Zeng and Hao Zhang and Feng Li and Jie Yang and Hongyang Li and Qing Jiang and Lei Zhang},
    year={2024},
    eprint={2401.14159},
    archivePrefix={arXiv},
    primaryClass={cs.CV}
}

@inproceedings{cherti_reproducible_2023,
	title = {Reproducible scaling laws for contrastive language-image learning},
	booktitle = {{IEEE}/{CVF} {Conference} on {Computer} {Vision} and {Pattern} {Recognition}},
	author = {Cherti, Mehdi and Beaumont, Romain and Wightman, Ross and Wortsman, Mitchell and Ilharco, Gabriel and Gordon, Cade and Schuhmann, Christoph and Schmidt, Ludwig and Jitsev, Jenia},
	year = {2023},
}

@article{balaji_ediff-i_2022,
  title = {{eDiff}-{I}: {Text}-to-{Image} {Diffusion} {Models} with an {Ensemble} of {Expert} {Denoisers}},
  author={Balaji, Yogesh and Nah, Seungjun and Huang, Xun and Vahdat, Arash and Song, Jiaming and Zhang, Qinsheng and Kreis, Karsten and Aittala, Miika and Aila, Timo and Laine, Samuli and others},
  journal={arXiv preprint arXiv:2211.01324},
  year={2022}
}

@inproceedings{abdal_image2stylegan_2019,
	title = {{Image2StyleGAN}: {How} to {Embed} {Images} {Into} the {StyleGAN} {Latent} {Space}?},
	booktitle = {{IEEE}/{CVF} {International} {Conference} on {Computer} {Vision}},
	author = {Abdal, Rameen and Qin, Yipeng and Wonka, Peter},
	year = {2019},
}

@article{jang_decordecomposition_2024,
  title = {{DECOR}:{Decomposition} and {Projection} of {Text} {Embeddings} for {Text}-to-{Image} {Customization}},
  author={Jang, Geonhui and Kim, Jin-Hwa and Park, Yong-Hyun and Kim, Junho and Lee, Gayoung and Jeong, Yonghyun},
  journal={arXiv preprint arXiv:2412.09169},
  year={2024}
}

@inproceedings{shah_ziplora_2024,
	title = {{ZipLoRA}: {Any} {Subject} in {Any} {Style} by {Effectively} {Merging} {LoRAs}},
	booktitle = {European {Conference} on {Computer} {Vision}},
	author = {Shah, Viraj and Ruiz, Nataniel and Cole, Forrester and Lu, Erika and Lazebnik, Svetlana and Li, Yuanzhen and Jampani, Varun},
	year = {2024},
}

@inproceedings{ruiz_hyperdreambooth_2024,
	title = {{HyperDreamBooth}: {HyperNetworks} for {Fast} {Personalization} of {Text}-to-{Image} {Models}},
	shorttitle = {{HyperDreamBooth}},
	booktitle = {{IEEE}/{CVF} {Conference} on {Computer} {Vision} and {Pattern} {Recognition}},
	author = {Ruiz, Nataniel and Li, Yuanzhen and Jampani, Varun and Wei, Wei and Hou, Tingbo and Pritch, Yael and Wadhwa, Neal and Rubinstein, Michael and Aberman, Kfir},
	year = {2024},
}

@inproceedings{lee_direct_2024,
	title = {Direct {Consistency} {Optimization} for {Compositional} {Text}-to-{Image} {Personalization}},
	booktitle = {Advances in {Neural} {Information} {Processing} {Systems}},
	author = {Lee, Kyungmin and Kwak, Sangkyung and Sohn, Kihyuk and Shin, Jinwoo},
	year = {2024},
}

@inproceedings{ramesh_zero-shot_2021,
  title = {Zero-{Shot} {Text}-to-{Image} {Generation}},
  author={Ramesh, Aditya and Pavlov, Mikhail and Goh, Gabriel and Gray, Scott and Voss, Chelsea and Radford, Alec and Chen, Mark and Sutskever, Ilya},
  booktitle={International Conference on Machine Learning},
  pages={8821--8831},
  year={2021},
  organization={PMLR}
}

@inproceedings{karras_analyzing_2020,
	title = {Analyzing and {Improving} the {Image} {Quality} of {StyleGAN}},
	booktitle = {{IEEE}/{CVF} {Conference} on {Computer} {Vision} and {Pattern} {Recognition}},
	author = {Karras, Tero and Laine, Samuli and Aittala, Miika and Hellsten, Janne and Lehtinen, Jaakko and Aila, Timo},
	year = {2020},
}

@inproceedings{karras_style-based_2019,
	title = {A {Style}-{Based} {Generator} {Architecture} for {Generative} {Adversarial} {Networks}},
	booktitle = {{IEEE}/{CVF} {Conference} on {Computer} {Vision} and {Pattern} {Recognition}},
	author = {Karras, Tero and Laine, Samuli and Aila, Timo},
	year = {2019},
}

@article{tov_designing_2021,
	title = {Designing an encoder for {StyleGAN} image manipulation},
	volume = {40},
	issn = {0730-0301},
	doi = {10.1145/3450626.3459838},
	number = {4},
	urldate = {2025-03-05},
	journal = {ACM Trans. Graph.},
	author = {Tov, Omer and Alaluf, Yuval and Nitzan, Yotam and Patashnik, Or and Cohen-Or, Daniel},
	year = {2021},
	pages = {133:1--133:14},
}

@inproceedings{richardson_encoding_2021,
	title = {Encoding in {Style}: a {StyleGAN} {Encoder} for {Image}-to-{Image} {Translation}},
	booktitle = {{IEEE}/{CVF} {Conference} on {Computer} {Vision} and {Pattern} {Recognition}},
	author = {Richardson, Elad and Alaluf, Yuval and Patashnik, Or and Nitzan, Yotam and Azar, Yaniv and Shapiro, Stav and Cohen-Or, Daniel},
	year = {2021},
}

@inproceedings{esser_scaling_2024,
  title = {Scaling {Rectified} {Flow} {Transformers} for {High}-{Resolution} {Image} {Synthesis}},
  author={Esser, Patrick and Kulal, Sumith and Blattmann, Andreas and Entezari, Rahim and M{\"u}ller, Jonas and Saini, Harry and Levi, Yam and Lorenz, Dominik and Sauer, Axel and Boesel, Frederic and others},
  booktitle={International Conference on Machine Learning},
  year={2024}
}

@inproceedings{houlsby_parameter-efficient_2019,
  title = {Parameter-{Efficient} {Transfer} {Learning} for {NLP}},
  author={Houlsby, Neil and Giurgiu, Andrei and Jastrzebski, Stanislaw and Morrone, Bruna and De Laroussilhe, Quentin and Gesmundo, Andrea and Attariyan, Mona and Gelly, Sylvain},
  booktitle={International Conference on Machine Learning},
  pages={2790--2799},
  year={2019},
  organization={PMLR}
}

@inproceedings{kumari_multi-concept_2023,
	title = {Multi-{Concept} {Customization} of {Text}-to-{Image} {Diffusion}},
	booktitle = {{IEEE} {Conference} on {Computer} {Vision} and {Pattern} {Recognition}},
	author = {Kumari, Nupur and Zhang, Bingliang and Zhang, Richard and Shechtman, Eli and Zhu, Jun-Yan},
	year = {2023},
}

@inproceedings{wei_elite_2023,
	title = {{ELITE}: {Encoding} {Visual} {Concepts} into {Textual} {Embeddings} for {Customized} {Text}-to-{Image} {Generation}},
	booktitle = {{IEEE}/{CVF} {International} {Conference} on {Computer} {Vision}},
	author = {Wei, Yuxiang and Zhang, Yabo and Ji, Zhilong and Bai, Jinfeng and Zhang, Lei and Zuo, Wangmeng},
	year = {2023},
}

@article{oquab_dinov2_2024,
	title = {{DINOv2}: {Learning} {Robust} {Visual} {Features} without {Supervision}},
	issn = {2835-8856},
	journal = {Transactions on Machine Learning Research},
	author = {Oquab, Maxime and Darcet, Timothée and Moutakanni, Théo and Vo, Huy and Szafraniec, Marc and Khalidov, Vasil and Fernandez, Pierre and Haziza, Daniel and Massa, Francisco and El-Nouby, Alaaeldin and Assran, Mahmoud and Ballas, Nicolas and Galuba, Wojciech and Howes, Russell and Huang, Po-Yao and Li, Shang-Wen and Misra, Ishan and Rabbat, Michael and Sharma, Vasu and Synnaeve, Gabriel and Xu, Hu and Jegou, Hervé and Mairal, Julien and Labatut, Patrick and Joulin, Armand and Bojanowski, Piotr},
	year = {2024},
}

@inproceedings{lin_evaluating_2024,
	title = {Evaluating {Text}-to-{Visual} {Generation} with {Image}-to-{Text} {Generation}},
	shorttitle = {{VQAScore}},
	booktitle = {European {Conference} on {Computer} {Vision}},
	author = {Lin, Zhiqiu and Pathak, Deepak and Li, Baiqi and Li, Jiayao and Xia, Xide and Neubig, Graham and Zhang, Pengchuan and Ramanan, Deva},
	year = {2024},
}

@inproceedings{li_textcraftor_2024,
	title = {{TextCraftor}: {Your} {Text} {Encoder} {Can} be {Image} {Quality} {Controller}},
	booktitle = {{IEEE}/{CVF} {Conference} on {Computer} {Vision} and {Pattern} {Recognition}},
	author = {Li, Yanyu and Liu, Xian and Kag, Anil and Hu, Ju and Idelbayev, Yerlan and Sagar, Dhritiman and Wang, Yanzhi and Tulyakov, Sergey and Ren, Jian},
	year = {2024},
}

@inproceedings{saharia_photorealistic_2022,
	title = {Photorealistic {Text}-to-{Image} {Diffusion} {Models} with {Deep} {Language} {Understanding}},
	booktitle = {Advances in {Neural} {Information} {Processing} {Systems}},
	author = {Saharia, Chitwan and Chan, William and Saxena, Saurabh and Li, Lala and Whang, Jay and Denton, Emily and Ghasemipour, Seyed Kamyar Seyed and Ayan, Burcu Karagol and Mahdavi, S. Sara and Lopes, Rapha Gontijo and Salimans, Tim and Ho, Jonathan and Fleet, David J. and Norouzi, Mohammad},
	year = {2022},
}

@inproceedings{hu_lora_2022,
	title = {{LoRA}: {Low}-{Rank} {Adaptation} of {Large} {Language} {Models}},
	booktitle = {International {Conference} on {Learning} {Representations}},
	author = {Hu, Edward J. and Shen, Yelong and Wallis, Phillip and Allen-Zhu, Zeyuan and Li, Yuanzhi and Wang, Shean and Wang, Lu and Chen, Weizhu},
	year = {2022},
}

@inproceedings{qiu_controlling_2023,
	title = {Controlling {Text}-to-{Image} {Diffusion} by {Orthogonal} {Finetuning}},
	booktitle = {Advances in {Neural} {Information} {Processing} {Systems}},
	author = {Qiu, Zeju and Liu, Weiyang and Feng, Haiwen and Xue, Yuxuan and Feng, Yao and Liu, Zhen and Zhang, Dan and Weller, Adrian and Schölkopf, Bernhard},
	year = {2023},
}

@article{alaluf_neural_2023,
	title = {A {Neural} {Space}-{Time} {Representation} for {Text}-to-{Image} {Personalization}},
	volume = {42},
	journal = {ACM Transactions on Graphics (TOG)},
	author = {Alaluf, Yuval and Richardson, Elad and Metzer, Gal and Cohen-Or, Daniel},
	year = {2023},
}

@inproceedings{ronneberger_u-net_2015,
	title = {U-{Net}: {Convolutional} {Networks} for {Biomedical} {Image} {Segmentation}},
	shorttitle = {U-{Net}},
	language = {en},
	booktitle = {International {Conference} on {Medical} {Image} {Computing} and {Computer}-{Assisted} {Intervention}},
	author = {Ronneberger, Olaf and Fischer, Philipp and Brox, Thomas},
	year = {2015},
}

@inproceedings{loshchilov_decoupled_2019,
	title = {Decoupled {Weight} {Decay} {Regularization}},
	booktitle = {International {Conference} on {Learning} {Representations}},
	author = {Loshchilov, Ilya and Hutter, Frank},
	year = {2019},
}

@inproceedings{sohn_styledrop_2023,
	title = {{StyleDrop}: {Text}-to-{Image} {Generation} in {Any} {Style}},
	booktitle = {Advances in {Neural} {Information} {Processing} {Systems}},
	author = {Sohn, Kihyuk and Ruiz, Nataniel and Lee, Kimin and Chin, Daniel Castro and Blok, Irina and Chang, Huiwen and Barber, Jarred and Jiang, Lu and Entis, Glenn and Li, Yuanzhen and Hao, Yuan and Essa, Irfan and Rubinstein, Michael and Krishnan, Dilip},
	year = {2023},
}

@inproceedings{ruiz_dreambooth_2023,
	title = {{DreamBooth}: {Fine} {Tuning} {Text}-to-{Image} {Diffusion} {Models} for {Subject}-{Driven} {Generation}},
	booktitle = {{IEEE}/{CVF} {Conference} on {Computer} {Vision} and {Pattern} {Recognition}},
	author = {Ruiz, Nataniel and Li, Yuanzhen and Jampani, Varun and Pritch, Yael and Rubinstein, Michael and Aberman, Kfir},
	year = {2023},
}

@article{ramesh_hierarchical_2022,
  title = {Hierarchical {Text}-{Conditional} {Image} {Generation} with {CLIP} {Latents}},
  author={Ramesh, Aditya and Dhariwal, Prafulla and Nichol, Alex and Chu, Casey and Chen, Mark},
  journal={arXiv preprint arXiv:2204.06125},
  year={2022}
}

@article{voynov_p_2023,
  title = {P+: {Extended} {Textual} {Conditioning} in {Text}-to-{Image} {Generation}},
  author={Voynov, Andrey and Chu, Qinghao and Cohen-Or, Daniel and Aberman, Kfir},
  journal={arXiv preprint arXiv:2303.09522},
  year={2023}
}

@inproceedings{sohl-dickstein_deep_2015,
	title = {Deep {Unsupervised} {Learning} using {Nonequilibrium} {Thermodynamics}},
	booktitle = {{International} {Conference} on {Machine} {Learning}},
	publisher = {PMLR},
	author = {Sohl-Dickstein, Jascha and Weiss, Eric A. and Maheswaranathan, Niru and Ganguli, Surya},
	year = {2015},
}

@inproceedings{rombach_high-resolution_2022,
	title = {High-{Resolution} {Image} {Synthesis} with {Latent} {Diffusion} {Models}},
	booktitle = {{IEEE/CVF} {Conference} on {Computer} {Vision} and {Pattern} {Recognition}},
	author = {Rombach, Robin and Blattmann, Andreas and Lorenz, Dominik and Esser, Patrick and Ommer, Björn},
	year = {2022},
}

@inproceedings{radford_learning_2021,
	title = {Learning {Transferable} {Visual} {Models} {From} {Natural} {Language} {Supervision}},
	booktitle = {{International} {Conference} on {Machine} {Learning}},
	publisher = {PMLR},
	author = {Radford, Alec and Kim, Jong Wook and Hallacy, Chris and Ramesh, Aditya and Goh, Gabriel and Agarwal, Sandhini and Sastry, Girish and Askell, Amanda and Mishkin, Pamela and Clark, Jack and Krueger, Gretchen and Sutskever, Ilya},
	year = {2021},
}

@inproceedings{podell_sdxl_2024,
	title = {{SDXL}: {Improving} {Latent} {Diffusion} {Models} for {High}-{Resolution} {Image} {Synthesis}},
	booktitle = {International {Conference} on {Learning} {Representations}},
	author = {Podell, Dustin and English, Zion and Lacey, Kyle and Blattmann, Andreas and Dockhorn, Tim and Müller, Jonas and Penna, Joe and Rombach, Robin},
	year = {2024},
}

@inproceedings{nichol_glide_2023,
	title = {{GLIDE}: {Towards} {Photorealistic} {Image} {Generation} and {Editing} with {Text}-{Guided} {Diffusion} {Models}},
	booktitle = {{International} {Conference} on {Machine} {Learning}},
	publisher = {PMLR},
	author = {Nichol, Alex and Dhariwal, Prafulla and Ramesh, Aditya and Shyam, Pranav and Mishkin, Pamela and McGrew, Bob and Sutskever, Ilya and Chen, Mark},
	year = {2023},
}

@inproceedings{li_few-shot_2020,
	title = {Few-shot {Image} {Generation} with {Elastic} {Weight} {Consolidation}},
	booktitle = {Advances in {Neural} {Information} {Processing} {Systems}},
	author = {Li, Yijun and Zhang, Richard and Lu, Jingwan (Cynthia) and Shechtman, Eli},
	year = {2020},
}

@inproceedings{li_blip-diffusion_2023,
	title = {{BLIP}-{Diffusion}: {Pre}-trained {Subject} {Representation} for {Controllable} {Text}-to-{Image} {Generation} and {Editing}},
	booktitle = {Advances in {Neural} {Information} {Processing} {Systems}},
	author = {Li, Dongxu and Li, Junnan and Hoi, Steven C. H.},
	year = {2023},
}

@inproceedings{gal_image_2023,
	title = {An {Image} is {Worth} {One} {Word}: {Personalizing} {Text}-to-{Image} {Generation} using {Textual} {Inversion}},
	booktitle = {International {Conference} on {Learning} {Representations}},
	author = {Gal, Rinon and Alaluf, Yuval and Atzmon, Yuval and Patashnik, Or and Bermano, Amit H. and Chechik, Gal and Cohen-Or, Daniel},
	year = {2023},
}

@inproceedings{chen_disenbooth_2024,
	title = {{DisenBooth}: {Identity}-{Preserving} {Disentangled} {Tuning} for {Subject}-{Driven} {Text}-to-{Image} {Generation}},
	booktitle = {International {Conference} on {Learning} {Representations}},
	author = {Chen, Hong and Zhang, Yipeng and Wu, Simin and Wang, Xin and Duan, Xuguang and Zhou, Yuwei and Zhu, Wenwu},
	year = {2024},
}

@inproceedings{goodfellow_generative_2014,
	title = {Generative {Adversarial} {Nets}},
	booktitle = {Advances in {Neural} {Information} {Processing} {Systems}},
	author = {Goodfellow, Ian J and Pouget-Abadie, Jean and Mirza, Mehdi and Xu, Bing and Warde-Farley, David and Ozair, Sherjil and Courville, Aaron and Bengio, Yoshua},
	year = {2014},
}

@inproceedings{li2024textcraftor,
  title={Textcraftor: Your text encoder can be image quality controller},
  author={Li, Yanyu and Liu, Xian and Kag, Anil and Hu, Ju and Idelbayev, Yerlan and Sagar, Dhritiman and Wang, Yanzhi and Tulyakov, Sergey and Ren, Jian},
  booktitle={Proceedings of the IEEE/CVF Conference on Computer Vision and Pattern Recognition},
  pages={7985--7995},
  year={2024}
}

@inproceedings{chen2024enhancing,
  title={Enhancing diffusion models with text-encoder reinforcement learning},
  author={Chen, Chaofeng and Wang, Annan and Wu, Haoning and Liao, Liang and Sun, Wenxiu and Yan, Qiong and Lin, Weisi},
  booktitle={European Conference on Computer Vision},
  year={2024},
}

@inproceedings{um2025minority,
  title={Minority-focused text-to-image generation via prompt optimization},
  author={Um, Soobin and Ye, Jong Chul},
  booktitle={{IEEE/CVF} {Conference} on {Computer} {Vision} and {Pattern} {Recognition}},
  pages={20926--20936},
  year={2025}
}

@article{kumar2025deft,
  title={DEFT: Decompositional Efficient Fine-Tuning for Text-to-Image Models},
  author={Kumar, Komal and Anwer, Rao Muhammad and Khan, Fahad Shahbaz and Khan, Salman and Laptev, Ivan and Cholakkal, Hisham},
  journal={arXiv preprint arXiv:2509.22793},
  year={2025}
}

@inproceedings{chen2024personalizing,
  title={PaRa: Personalizing Text-to-Image Diffusion via Parameter Rank Reduction},
  author={Shangyu Chen and Zizheng Pan and Jianfei Cai and Dinh Phung},
  booktitle={International Conference on Learning Representations},
  year={2025},
}

@article{na2025diffusion,
  title={Diffusion Adaptive Text Embedding for Text-to-Image Diffusion Models},
  author={Na, Byeonghu and Park, Minsang and Sim, Gyuwon and Shin, Donghyeok and Bae, HeeSun and Kang, Mina and Kwon, Se Jung and Kang, Wanmo and Moon, Il-Chul},
  journal={arXiv preprint arXiv:2510.23974},
  year={2025}
}

@article{chung2023prompt,
  title={Prompt-tuning latent diffusion models for inverse problems},
  author={Chung, Hyungjin and Ye, Jong Chul and Milanfar, Peyman and Delbracio, Mauricio},
  journal={arXiv preprint arXiv:2310.01110},
  year={2023}
}
\bibliographystyle{tmlr}

%%%%%%%%%% APPENDIX %%%%%%%%%%
\clearpage
\appendix
\section*{Appendix}

\section{User Study Details}
\label{sec:app:user_study}
%------------------------------------------
\begin{figure*}[ht!]
    \centering
    \includegraphics[width=0.7\linewidth]{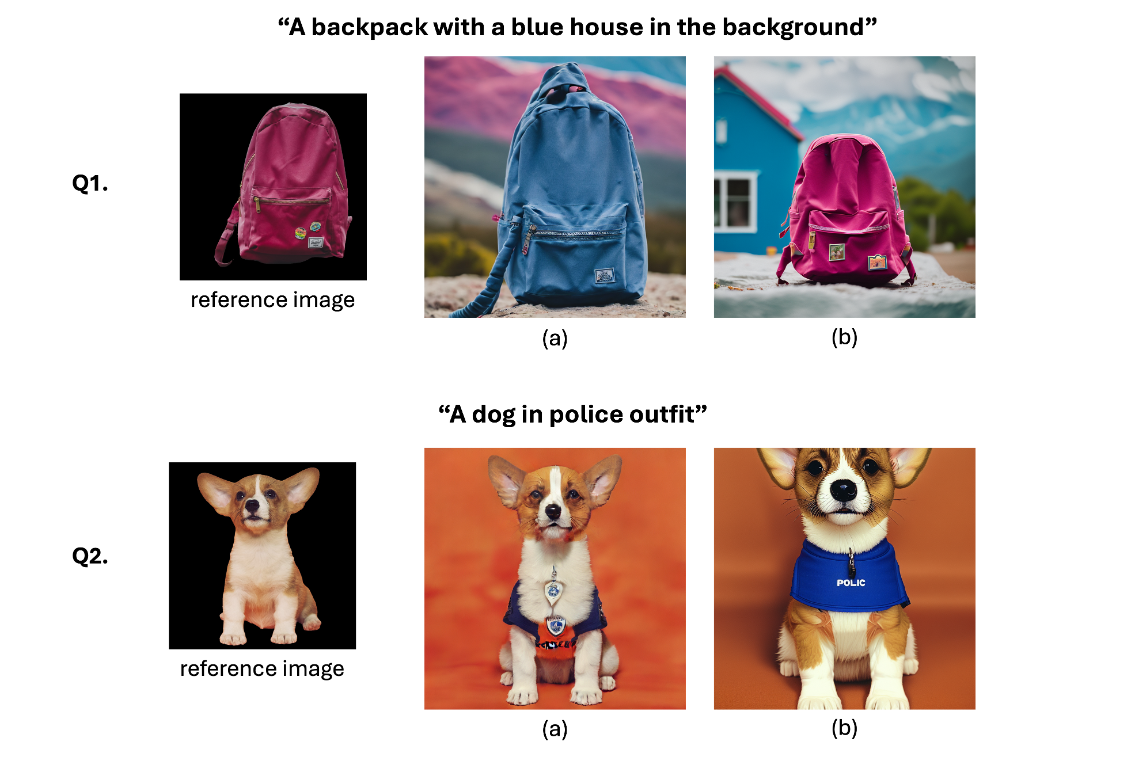}
    \caption{\textbf{User study template.} Each participant is asked to choose an image that best meets two criteria: subject fidelity and text prompt fidelity. We asked 50 participants to answer 30 questions each on Amazon Mechanical Turk and show the win-rate.}
    \label{app:us}
\end{figure*}
%------------------------------------------
To evaluate real-world user preferences for various personalization methods, we conducted a user study using Amazon Mechanical Turk. We recruited 50 participants, each of whom answered 30 questions. Participants were instructed to select the image that best satisfied two criteria simultaneously: (1) resembling the given subject and (2) adhering to the provided prompt.
For each question, users compared two images: one generated by our method and the other by a prior work. The prior methods used for comparison included Custom Diffusion~\citep{kumari_multi-concept_2023}, NeTI~\citep{alaluf_neural_2023}, and DisenBooth~\citep{chen_disenbooth_2024}, which represent the most recent advances in this domain. Example questions are provided in Fig.~\ref{app:us}.

To ensure a fair comparison, output images were randomly chosen (random seed and prompt) from the DreamBooth~\cite{ruiz_dreambooth_2023} prompt set. 
When comparing two images (as shown in Fig.~\ref{fig:userstudy} options (a) and (b)) we used the same random seed. Furthermore, to avoid position bias, the image options were presented to participants in a randomly shuffled order. The win rates for each method were calculated and are summarized in Fig.~\ref{fig:userstudy} of the main paper.

\section{Comparison with Existing Works}
\label{sec:app:comparison}
%------------------------------------------
\begin{figure*}[h!]
    \centering
    \includegraphics[width=0.65\linewidth]{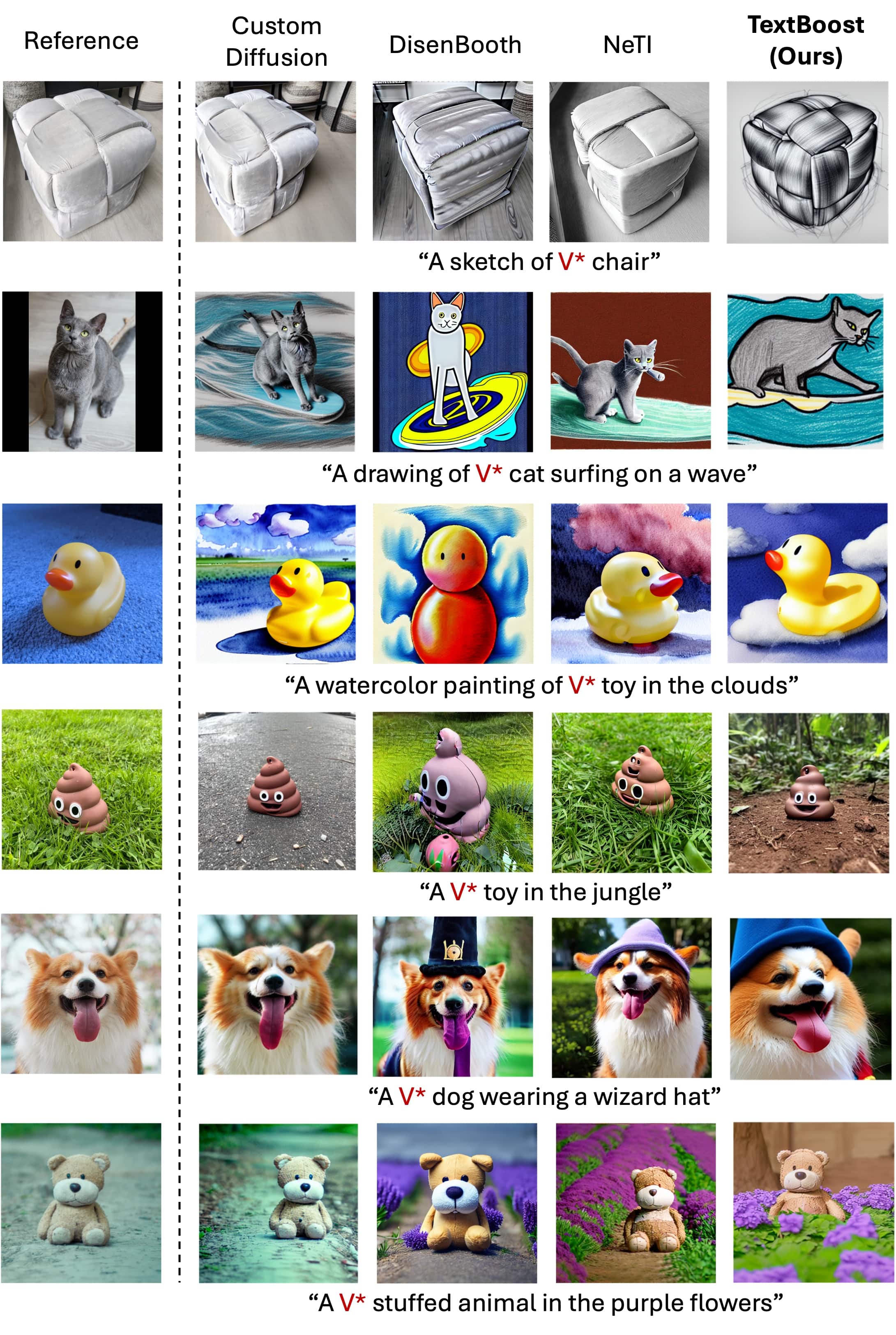}
    \caption{\textbf{More qualitative results.} We provide additional qualitative examples in comparison with recent personalization works. We used Stable Diffusion v1.5 as the baseline T2I model.}
    \label{app:cdtb}
\end{figure*}
%------------------------------------------
\subsection{Positioning of our work among text-side adaptation methods}
Recent work has explored the text-side of diffusion models in several different ways. 
Among methods that \emph{fine-tune the text encoder}, TextCraftor~\citep{li2024textcraftor} and TexForce~\citep{chen2024enhancing} show that adapting the text encoder can improve general image quality and text--image alignment, but they are not designed for instance-specific one-shot personalization. 
In contrast, PaRa~\citep{chen2024personalizing} and DEFT~\citep{kumar2025deft} are efficient personalization/adaptation frameworks that constrain or decompose model updates, but they are not specifically text-encoder-centered and do not address preserving the causal semantics of the text encoder. 
Meanwhile, a line of work operates on the \emph{text conditioning}: P2L~\citep{chung2023prompt} optimizes text embeddings for inverse problems, while DATE~\citep{na2025diffusion} and MinorityPrompt~\citep{um2025minority} dynamically adapt text embeddings or prompts during sampling to improve minority-sample generation. 
Compared with these works, our method targets \textbf{one-shot personalization} from a single reference image by selectively fine-tuning the text encoder, introducing Causality-Preserved Adaptation (CPA) to preserve the embeddings of tokens before \vstar, and further enhancing conditioning with layer-wise embeddings injected before each cross-attention layer.

\subsection{More experiments for comparison}
Here, we present more qualitative results compared to several recent works, as illustrated in Fig.~\ref{app:cdtb}.

For our baseline text-to-image (T2I) model, we employed Stable Diffusion v1.5. It is important to note that the results from Stable Diffusion v2.1-base and v2.1 are discussed in the main paper. Our method demonstrates a superior ability to generate images that closely align with the subjects of the reference images, significantly outperforming other approaches in capturing the essence of the given text prompts.

\begin{figure*}[h!]
    % \centering
    \includegraphics[width=0.95\linewidth]{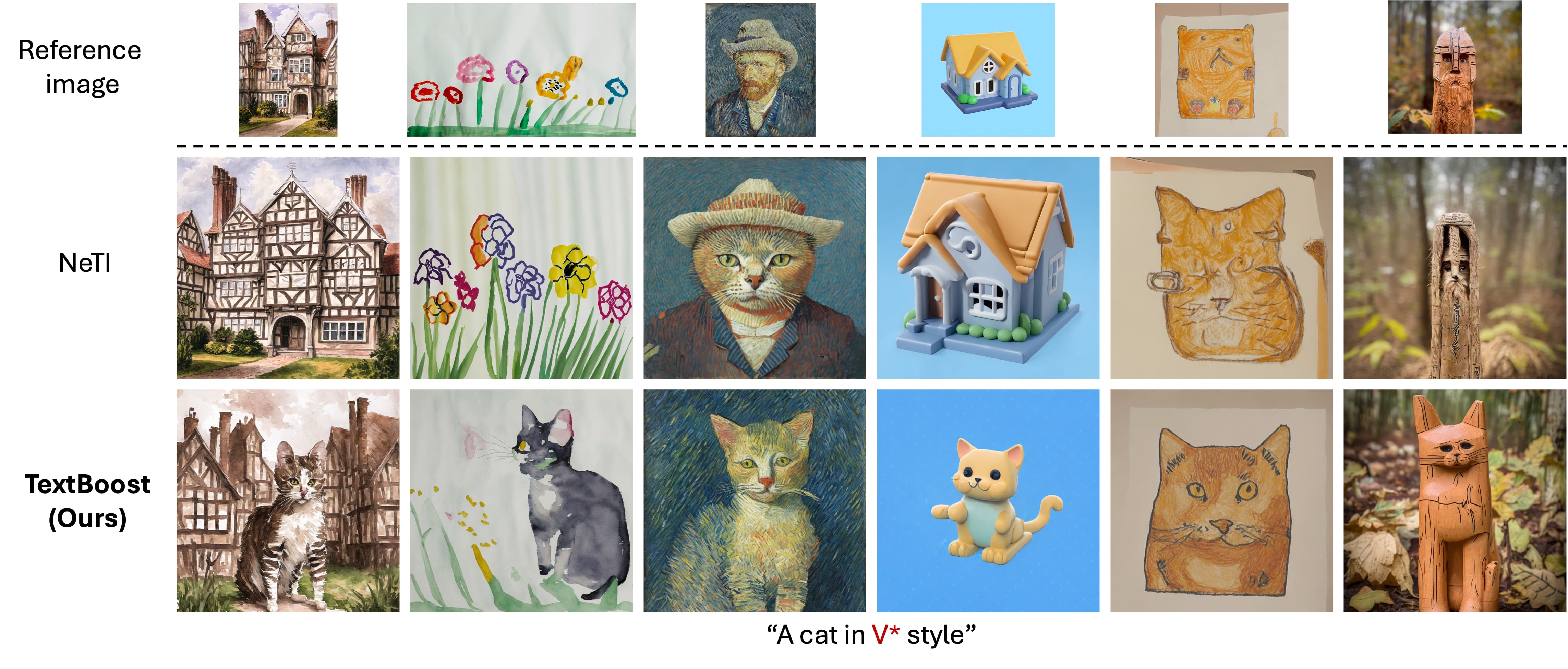}
    \caption{\textbf{Stylization.} We provide style personalization results of NeTI and compare with our \tb.}
    \label{fig:neti-style}
\end{figure*}

In the context of style personalization (i.e., stylization), we compared our approach with DisenBooth, as detailed in the main paper. Additionally, we explored stylization using NeTI~\citep{alaluf_neural_2023}, though these results were not included in the main paper due to NeTI's tendency to rapidly overfit. This overfitting results in the memorization of style, often at the expense of properly reflecting the input text prompts. A comparison with NeTI is provided in Fig.~\ref{fig:neti-style}.

\noindent \textbf{Implementation Details.}
We fine-tuned the baseline models following the experimental settings outlined in their respective original papers. However, since our study addresses the one-shot scenario, where only a single reference image is provided as training input, we decreased the training steps to mitigate overfitting in existing methods. 

For Stable Diffusion v1.5, we set the following training steps: Custom Diffusion~\citep{kumari_multi-concept_2023}: 250 steps, NeTI~\citep{alaluf_neural_2023}: 500 steps (without bypass), and DisenBooth~\citep{chen_disenbooth_2024}: 1,000 steps.

For style personalization, we used Stable Diffusion v2.1. We adjusted the batch size and training steps only if explicitly specified in the original paper; otherwise, we left them unchanged.

\section{Additional Results of \tb}
\label{sec:app:more_results}

\subsection{More results}
\begin{figure*}[h!]
    \centering
    \includegraphics[width=0.9\linewidth]{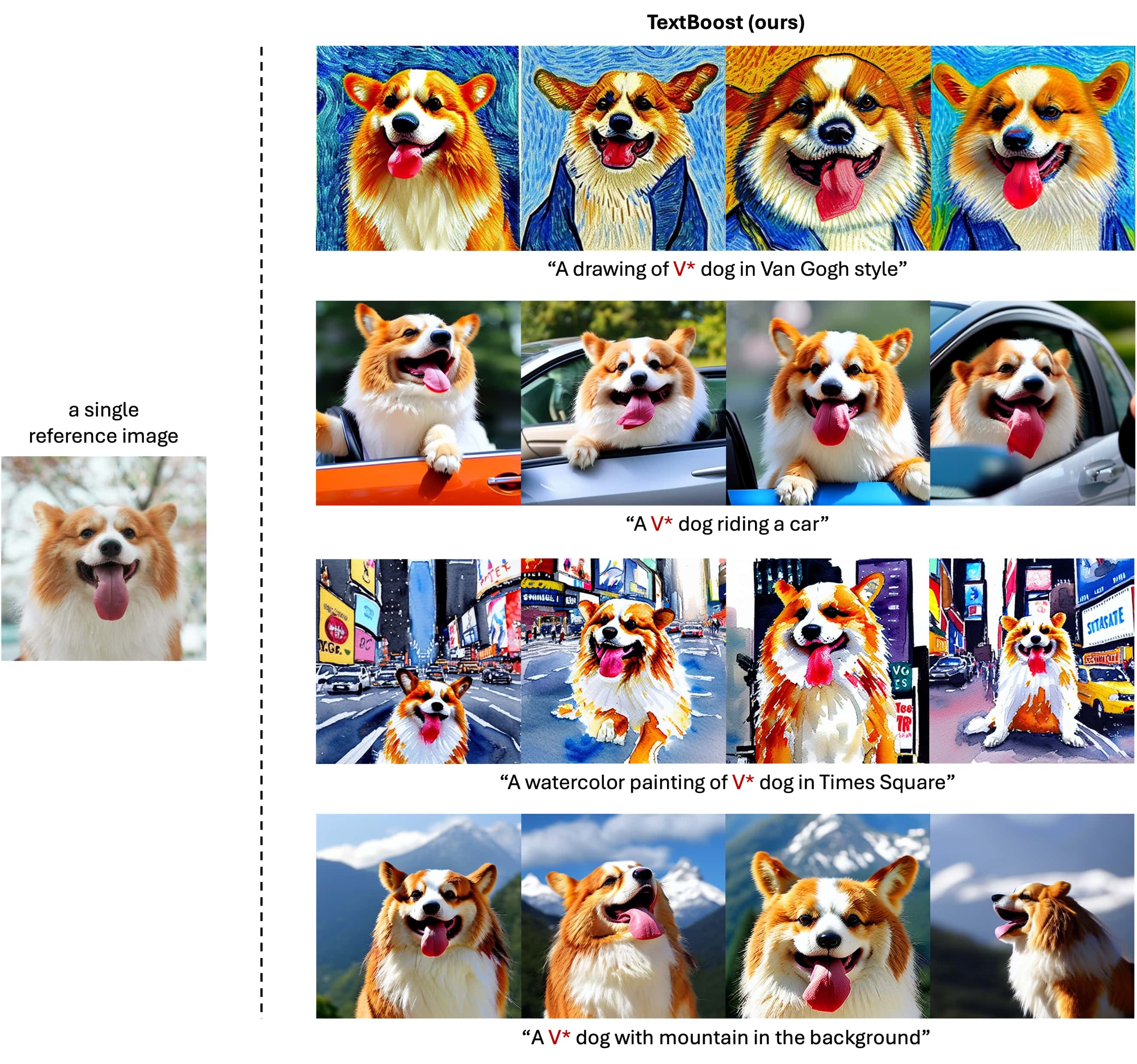}
    \caption{\textbf{More qualitative results of our \tb (dog).}}
    \label{app:ours1}
\end{figure*}

\begin{figure*}[h!]
    \centering
    \includegraphics[width=0.9\linewidth]{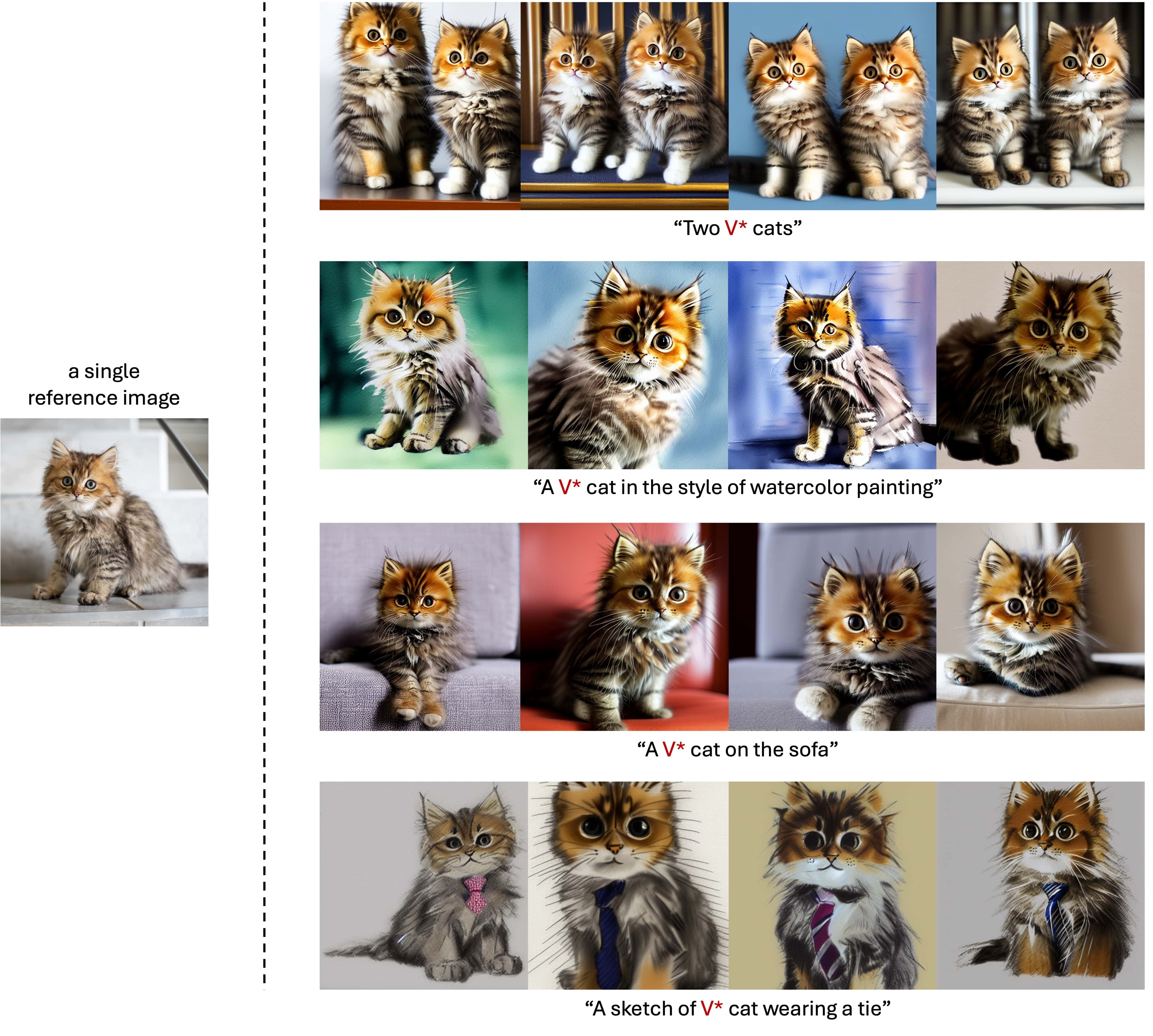}
    \caption{\textbf{More qualitative results of our \tb (cat).}}
    \label{app:ours2}
\end{figure*}

\begin{figure*}[h!]
    \centering
    \includegraphics[width=0.9\linewidth]{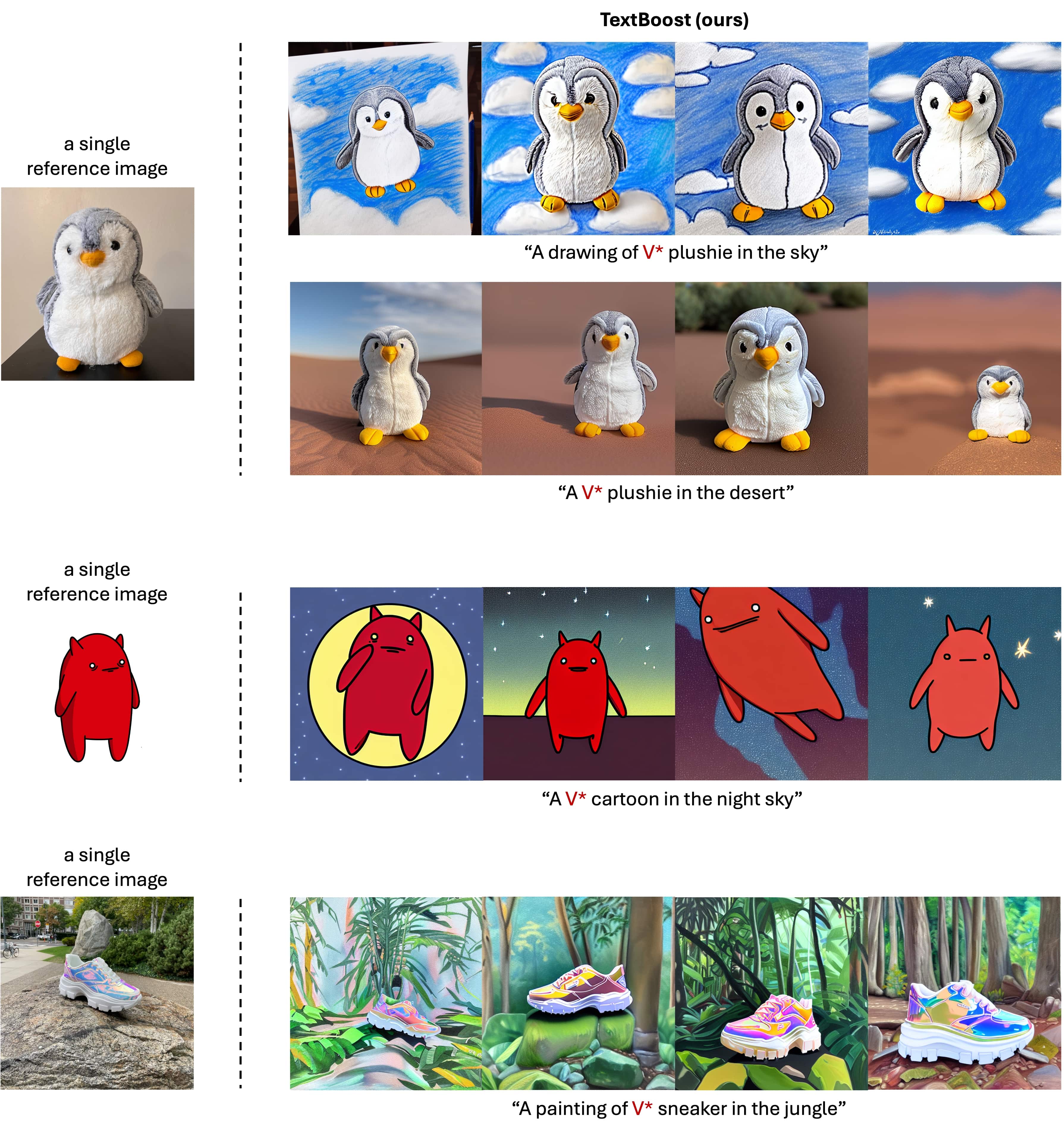}
    \caption{\textbf{More qualitative results of our \tb (several subjects).}}
    \label{app:ours3}
\end{figure*}

\begin{figure*}[h!]
    \centering
    \includegraphics[width=0.4\linewidth]{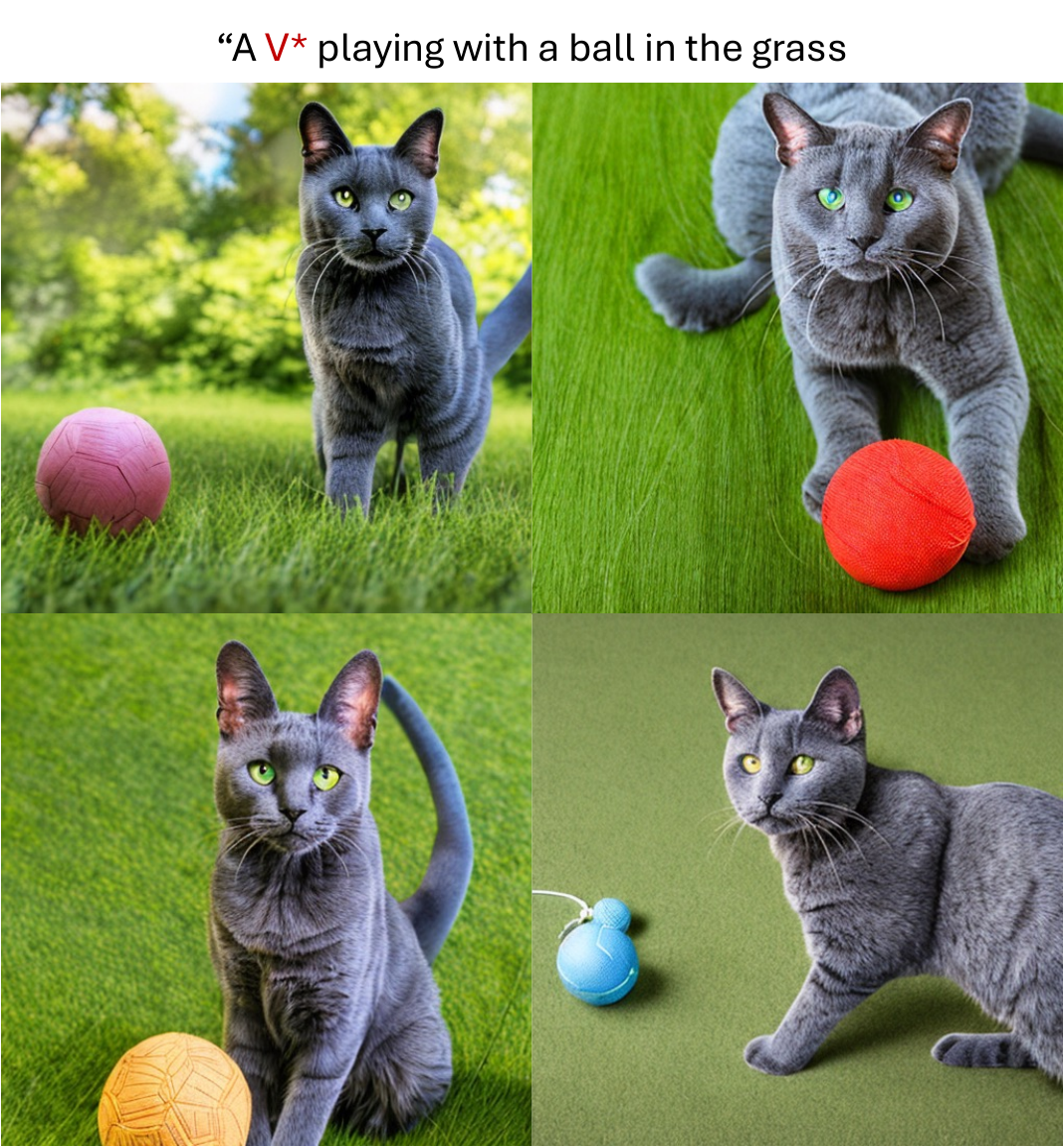}
    \caption{\textbf{Diversity results.} We show four generated outputs of our method with the same input text prompt. Our method, \tb, consistently generates images with high diversity.}
    \label{fig:supp_div}
\end{figure*}

We present additional qualitative results of our method in Figures \ref{app:ours1}, \ref{app:ours2}, \ref{app:ours3}, and \ref{fig:supp_div}.
To demonstrate its ability to produce diverse outputs, we include four generated images for each text prompt.
As shown in the figures, \tb generates diverse images from a single reference image for a given prompt. Furthermore, it consistently produces high-quality results, offering creative control through imaginative prompts with high diversity.

Furthermore, we present an ablation study (Tab.~\ref{app:tab:sdxl}) on fine-tuning the SDXL text encoders. Although fine-tuning both encoders achieves strong performance, we opt to fine-tune only the larger OpenCLIP encoder, which offers a good balance between performance and computational efficiency.

\begin{table}[t]
\centering
\caption{\textbf{Effect of fine-tuning different text encoders in SDXL.}
We compare personalization performance when fine-tuning only the CLIP ViT-L encoder, only the OpenCLIP encoder, or both encoders. 
Fine-tuning both encoders improves subject fidelity (DINO), while fine-tuning only CLIP preserves prompt adherence (VQA) better.}
\label{app:tab:sdxl}
\begin{tabular}{lcc}
\toprule
\textbf{Fine-tuned encoder} & \textbf{DINO} $\uparrow$ & \textbf{VQA} $\uparrow$ \\
\midrule
CLIP ViT-L only & 0.515 & 0.839 \\
OpenCLIP only & 0.565 & 0.763 \\
Both & 0.615 & 0.690 \\
\bottomrule
\end{tabular}
\end{table}

\subsection{Effect of CPA}
In Fig.~\ref{fig:causal}, we demonstrate that the proposed CPA is designed to preserve the prefix tokens that appear before \vstar. Although masking is intended to enforce this behavior, we further verify it empirically by measuring the cosine similarity between the prefix token representations produced by a base text encoder and those produced by \tb (which applies CPA). The mean similarity (across all subjects) between the tokens of \tb and the base text encoder is 1.00, indicating that the prefix representations are effectively preserved.

%------------------------------------------
\begin{figure}[t!]
    \centering
    \includegraphics[width=0.9\linewidth]{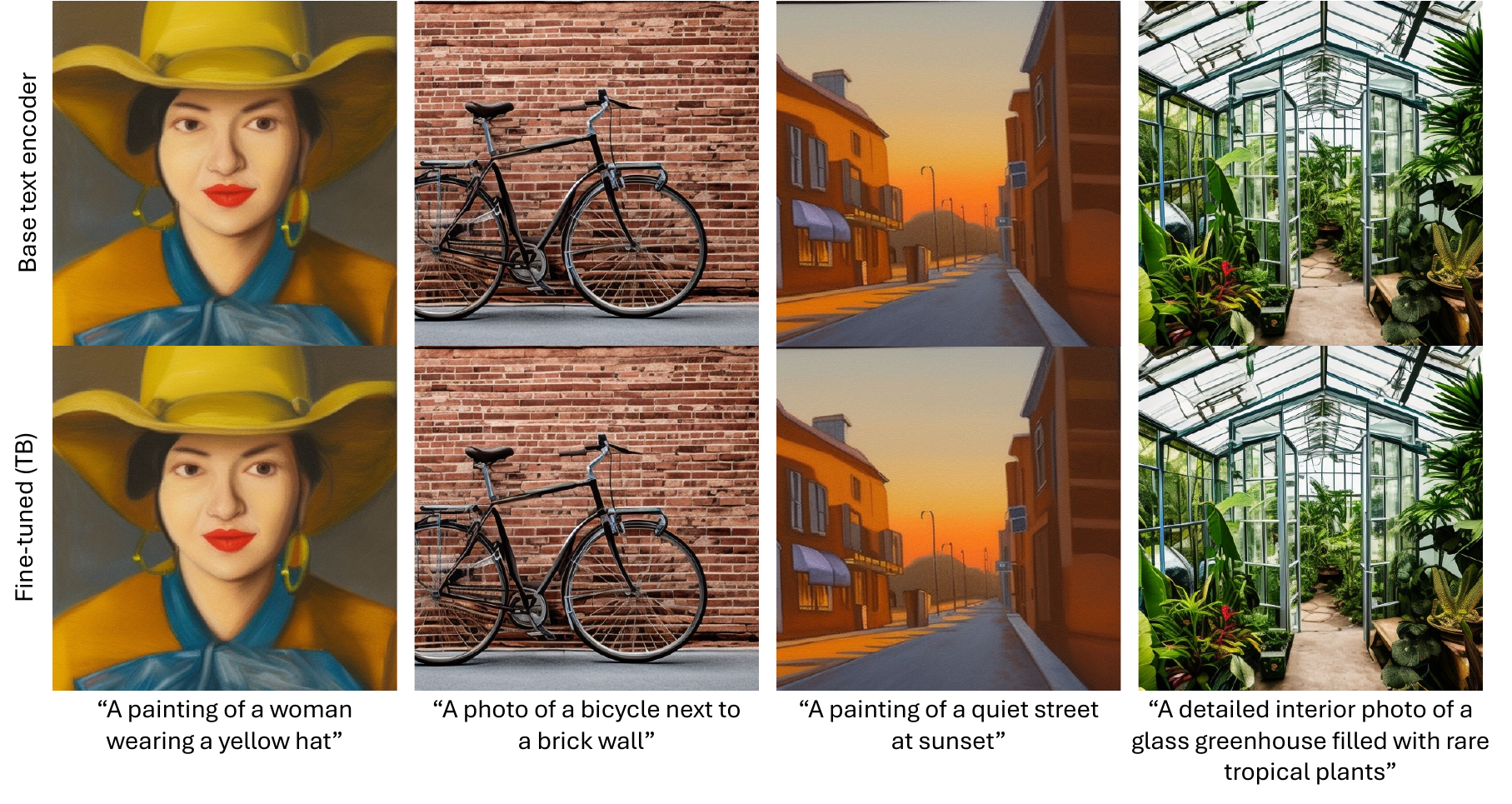}
    \caption{\textbf{Preserving text encoder behavior.} We compare images generated using the original (non-fine-tuned) text encoder and our \tb. The results show that the original token representations are preserved, as the generated images exhibit no noticeable visual differences.}
    \label{app:fig:no_vstar}
\end{figure}
%------------------------------------------

In addition, the behavior of the original text encoder remains unchanged for prompts that do not include \vstar. 
This is validated in the qualitative examples of Fig.~\ref{app:fig:no_vstar}, where the generated images remain visually identical.

%------------------------------------------
\begin{figure}[t!]
    \centering
    \includegraphics[width=\linewidth]{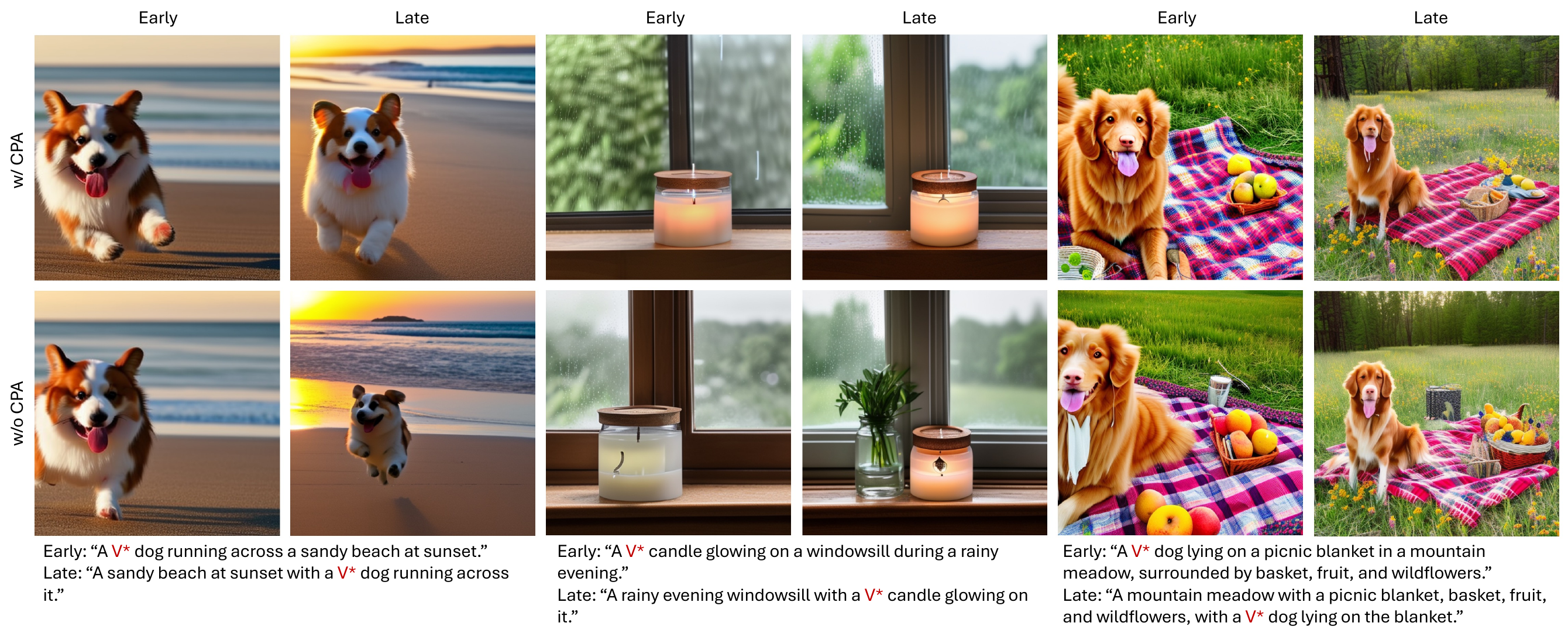}
    \caption{\textbf{Results with varying \vstar position.}
    We compare generations with and without CPA when the position of \vstar in the prompt changes. 
    Since CPA preserves the representations of tokens preceding \vstar, its effect becomes more visible when \vstar appears later in the sentence, where a longer prefix is preserved.}
    \label{app:fig:v_position}
\end{figure}
%------------------------------------------

Furthermore, we analyze the effect of CPA under different positions of the \vstar token in the prompt. Since CPA preserves the representations of tokens preceding \vstar, the impact of CPA becomes more noticeable when \vstar appears later in the sentence, where a longer prefix must be preserved. As illustrated in Fig.~\ref{app:fig:v_position}, CPA tends to better maintain prompt fidelity in such cases.

\begin{figure}[ht!]
    \centering
    \includegraphics[width=0.5\linewidth]{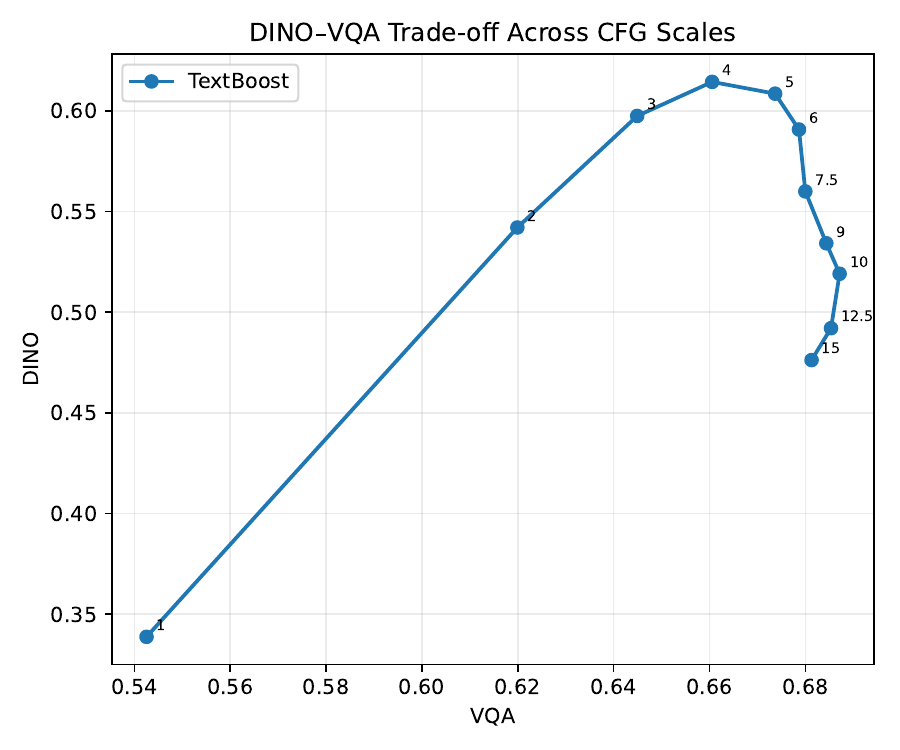}
    \caption{\textbf{DINO-VQA trade-off under a sweep of classifier-free guidance (CFG) scales.} Each point corresponds to one CFG value, so the plot shows how the operating point of \tb changes as the conditioning strength varies.}
    \label{fig:cfg_sweep}
\end{figure}
% \subsection{Effect of classifier-free guidance scale}
\subsection{Trade-off across conditioning strength.}
We performed a DINO-VQA trade-off experiment (Fig.~\ref{fig:cfg_sweep}) by sweeping the classifier-free guidance (CFG) scale at inference time. 
Since CFG directly controls the strength of conditioning, varying it traces an operating curve for each method rather than reporting only a single best-tuned setting. This visualization, therefore, shows how the quality of personalization changes as conditioning is strengthened or relaxed.

\section{Additional Results of \tbpp}
\begin{figure}[t]
    \centering
    \includegraphics[width=0.9\linewidth]{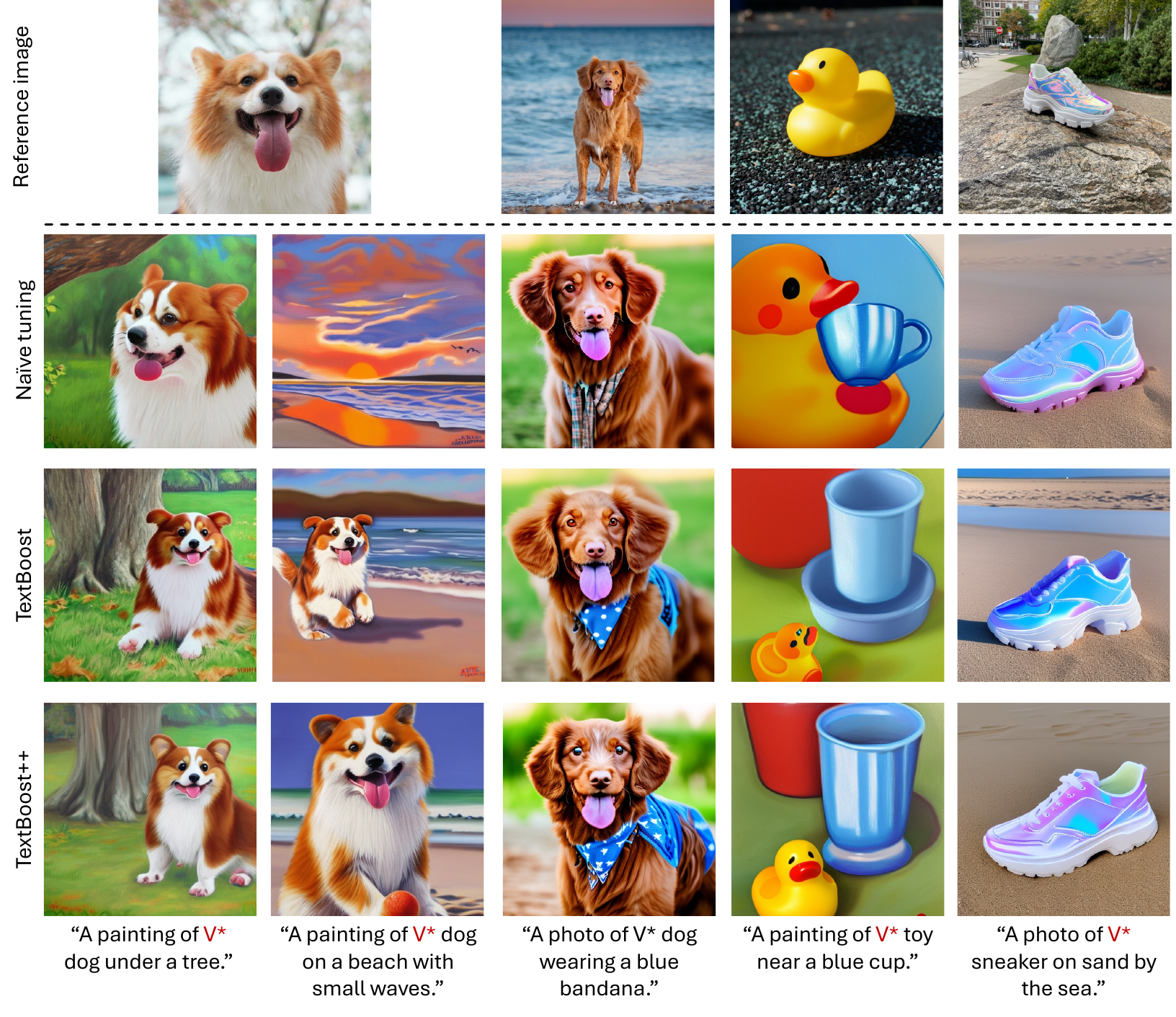}
    \caption{\textbf{Comprehensive comparison.} We compare naive text encoder fine-tuning, \tb, and \tbpp. Across the generated images, tb improves prompt adherence compared to naive tuning, while \tbpp improves subject fidelity.}
    \label{app:fig:comp_comp}
\end{figure}

To provide a comprehensive comparison of text encoder tuning approaches (naive tuning, \tb, and \tbpp), we present generated images with diverse subjects and prompts in Fig.~\ref{app:fig:comp_comp}. As shown in the figure, \tb improves text prompt fidelity compared to naive tuning. Furthermore, \tbpp achieves better subject fidelity than tb, which is consistent with the results reported in Tab.~\ref{tab:clipscore}.
%------------------------------------------
%------------------------------------------

% \begin{table}[t]
% \centering
% \begin{tabular}{lcc}
% \toprule
% Layer idx & DINO $\uparrow$ & VQA $\uparrow$ \\
% \midrule
% 0     & 0.573 & 0.710 \\
% 4     & 0.576 & 0.709 \\
% 8     & 0.573 & 0.712 \\
% 12    & 0.582 & 0.709 \\
% 16    & 0.596 & 0.708 \\
% \midrule
% All   & 0.583 & 0.727 \\
% \bottomrule
% \end{tabular}
% \caption{\blue{\textbf{Layer ablation of TextBoost++} Performance when applying LoRA to different layers of the text encoder.}}
% \label{app:tab:ab_blocks}
% \end{table}

%------------------------------------------
%------------------------------------------

%------------------------------------------
%------------------------------------------

%------------------------------------------
%------------------------------------------

%------------------------------------------
%------------------------------------------

\clearpage
\section{Experiments on Human Faces}
\label{sec:app:face}
%------------------------------------------
\begin{figure*}[h!]
    \centering
    \includegraphics[width=0.9\linewidth]{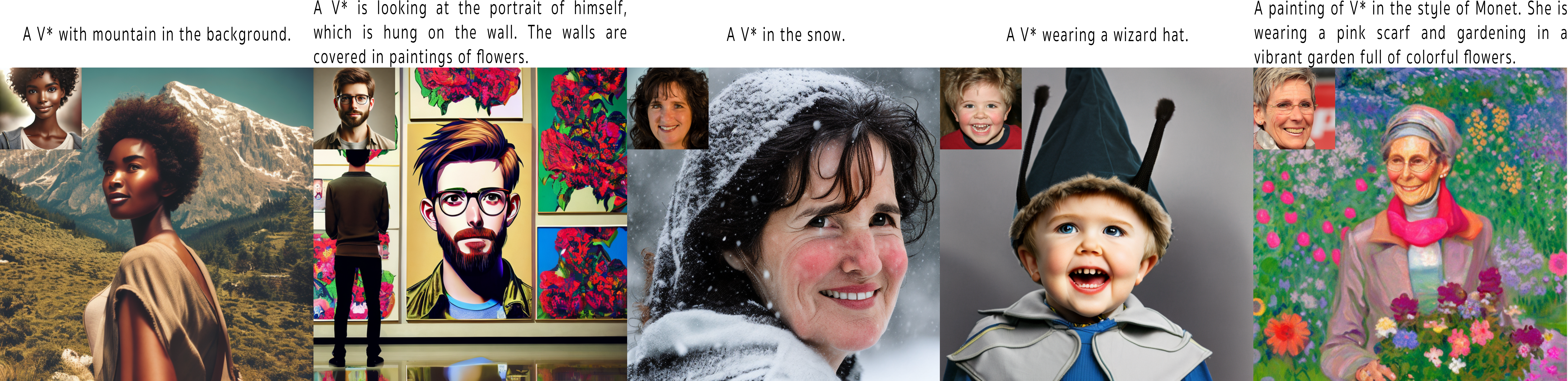}
    \caption{\textbf{Qualitative results on human faces.} We use 3 random faces from FFHQ and 2 generated images from DALL-E. We then use a single reference face image to train our \tb. The reference face image is located at the upper-left corner of each generated output.}
    \label{app:face}
\end{figure*}
%------------------------------------------
To demonstrate the versatility of \tb on a broader range of reference images, we conduct experiments using facial images. To avoid using celebrity images, we select three random faces from FFHQ and two synthetic faces generated by DALL-E. The results, shown in Fig.~\ref{app:face}, highlight that even with complex captions, \tb produces high-fidelity images with strong text-image alignment, demonstrating clear advantages of our approach.

\end{document}